%% file: acl_latex.tex
\pdfoutput=1
\documentclass[11pt]{article}

\usepackage[final]{acl}

\usepackage{times}
\usepackage{latexsym}

\usepackage[T1]{fontenc}
\usepackage[utf8]{inputenc}

\usepackage{microtype}

\usepackage{inconsolata}

\usepackage{graphicx}  % 移除重复的 graphicx
\usepackage{multirow}
\usepackage{booktabs}  % 移除重复的 booktabs
\usepackage[export]{adjustbox} 
\usepackage{longtable}
\usepackage{wrapfig,lipsum,tabularx}
\usepackage{makecell}
\usepackage{caption}

\usepackage{fontawesome5}
\usepackage{pifont}

\usepackage{url}

\usepackage{hyperref}

\definecolor{1}{HTML}{A0522D}
\definecolor{2}{HTML}{DEB887}
\definecolor{3}{HTML}{DBE0ED}
\definecolor{4}{HTML}{87B5B2}
\definecolor{5}{HTML}{F4CEB4}
\definecolor{6}{HTML}{EEC79F}

\title{Towards Dynamic Theory of Mind: Evaluating LLM Adaptation to Temporal Evolution of Human States}

\author{Yang Xiao$^{1}$\footnotemark[1] \quad Jiashuo Wang$^{1}$\footnotemark[1] \quad Qiancheng Xu$^{1}$ \quad Changhe Song$^{1}$ \\
\textbf{Chunpu Xu}$^{1}$ \quad \textbf{Yi Cheng}$^{1}$ \quad \textbf{Wenjie Li}$^{1}$\footnotemark[2] \quad \textbf{Pengfei Liu}$^{2}$\footnotemark[2] \\
$^{1}$The Hong Kong Polytechnic University \quad $^{2}$Shanghai Jiao Tong University \\
\texttt{yang-alan.xiao@connect.polyu.hk} \quad \texttt{csjwang@comp.polyu.edu.hk} 
}

\begin{document}
\maketitle
{
\renewcommand{\thefootnote}{\fnsymbol{footnote}}
\footnotetext[1]{Equal contribution.}
\footnotetext[2]{Corresponding authors.}
}
\begin{abstract}
\input{section/abstract}
\end{abstract}

\begin{figure}[!ht]
    \centering
    \includegraphics[width=\hsize]{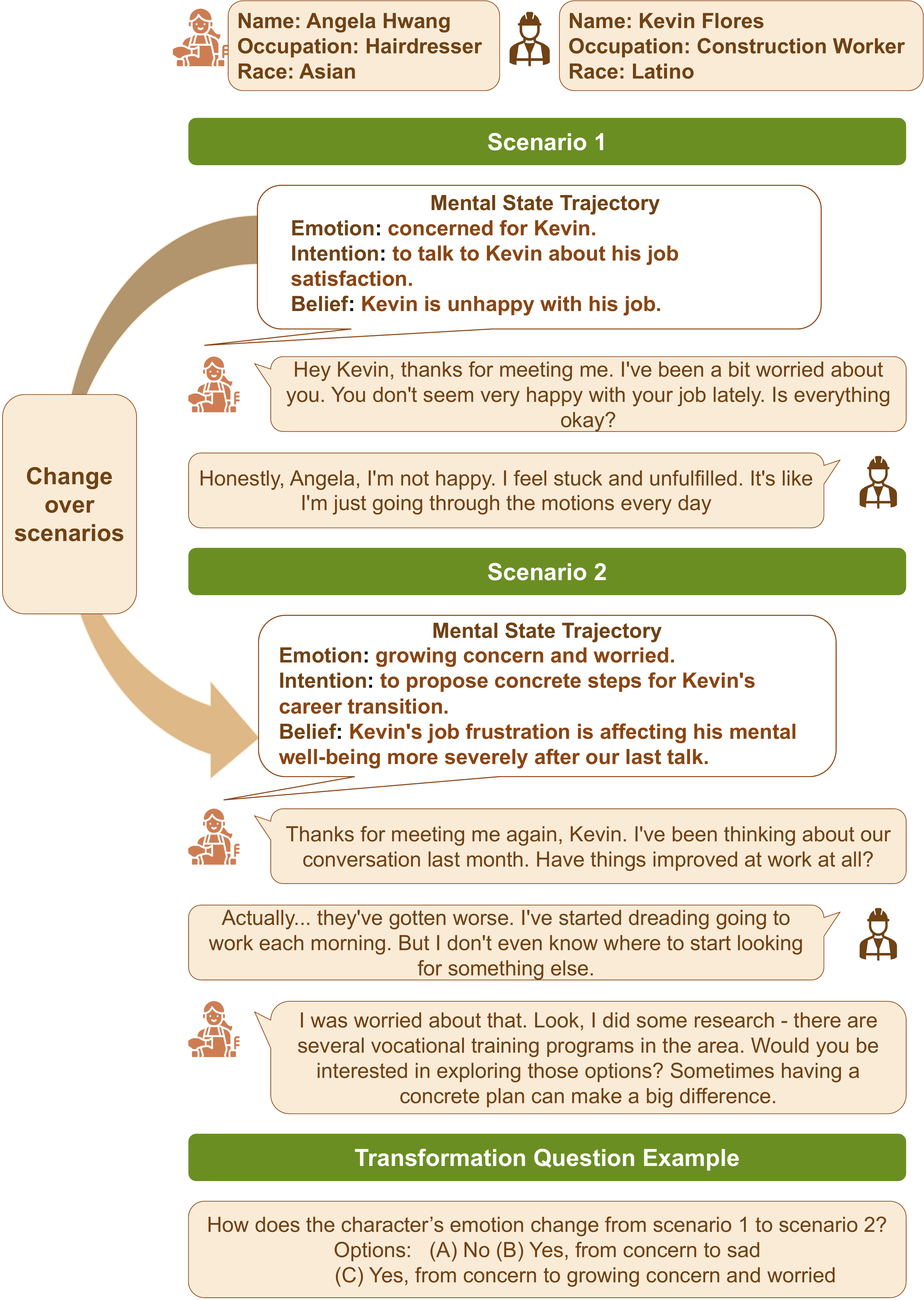}
    \caption{A simplified illustration showing the mental state trajectory's change over scenarios and the transformation question example to probe LLM's ability to adapt to the change.}
    \vspace{-10pt} 
    \label{fig:scenario1}
\end{figure}

\input{section/introduction}

\input{section/relatedwork}

\input{section/evaluation2construction}

\input{section/experiments}

\input{section/discussion}

\newpage
\bibliography{custom}
\clearpage
\appendix
\input{section/appendix}

\end{document}

%% file: section/abstract.tex
As Large Language Models (LLMs) increasingly participate in human-AI interactions, evaluating their Theory of Mind (ToM) capabilities - particularly their ability to track dynamic mental states - becomes crucial. While existing benchmarks assess basic ToM abilities, they predominantly focus on static snapshots of mental states, overlooking the temporal evolution that characterizes real-world social interactions. We present \textsc{DynToM}, a novel benchmark specifically designed to evaluate LLMs' ability to understand and track the temporal progression of mental states across interconnected scenarios. Through a systematic four-step framework, we generate 1,100 social contexts encompassing 5,500 scenarios and 78,100 questions, each validated for realism and quality. Our comprehensive evaluation of ten state-of-the-art LLMs reveals that their average performance underperforms humans by 44.7\%, with performance degrading significantly when tracking and reasoning about the shift of mental states. This performance gap highlights fundamental limitations in current LLMs' ability to model the dynamic nature of human mental states.
\footnote{\textsc{DynToM} are available at \href{https://github.com/GAIR-NLP/DynToM}{GitHub} and \href{https://huggingface.co/datasets/YangXiao-nlp/DynToM}{HuggingFace}.}

% \footnote{\url{https://anonymous.4open.science/r/ToMValley-ICLR/README.md}} 
% Our analysis provides detailed insights into specific challenges LLMs face in temporal reasoning about mental states, offering directions for improving their social understanding capabilities.

%% file: section/introduction.tex
\section{Introduction}

Theory of Mind (ToM) - the ability to understand and reason about others' mental states - is fundamental to human social interaction \citep{premack1978does,turner1988theory}. As Large Language Models (LLMs) increasingly engage in human-AI interactions, their capability to track and understand the dynamic nature of human mental states becomes crucial. While existing research has evaluated LLMs' ToM capabilities, these evaluations often overlook a critical aspect: the temporal evolution of mental states in real-world social contexts.

Current ToM evaluations of LLMs, including benchmarks like SocialIQA \citep{turner1988theory}, BigToM \citep{gandhi2023understanding}, and TOMBENCH \citep{chen-etal-2024-tombench}, predominantly focus on static snapshots of mental states in isolated scenarios. These works primarily focus on static evaluations, whereas our work presents a novel approach to capture the continuous change of mental states across multiple interconnected scenarios - a crucial aspect of real-world social interactions that has not been systematically evaluated in previous work. This temporal dimension is essential for understanding LLMs' true capabilities in real-world social interactions, where mental states constantly shift and evolve in response to ongoing social dynamics. For instance, LLMs are expected to understand and reason about the shift of user mental states in support conversations to better help users \citep{liu2024compeer,wang2024towards}.

To address this challenge, we introduce \textsc{DynToM} (Dynamic Theory of Mind), a novel benchmark designed specifically to evaluate LLMs' ability to track and understand the temporal evolution of mental states, as shown in Figure \ref{fig:scenario1}. Our benchmark is constructed through a systematic process: (1) social context construction, including social location, character profiles, and relationships; (2) mental state trajectory design across multiple scenarios; (3) scenario generation with natural dialogue; and (4) question formulation targeting temporal understanding of mental states. Each generated scenario and question undergoes rigorous human validation to ensure quality and realism. \textsc{DynToM} captures mental state dynamics through continuous social scenarios while incorporating real-world elements such as rich social contexts.

Our benchmark comprises 1,100 social contexts featuring 2,200 characters across 261 social locations, 5,500 social scenarios, and 78,100 multiple-choice questions. Through a comprehensive evaluation of ten representative LLMs, including GPT-4 series \citep{achiam2023gpt}, Llama 3 series \citep{dubey2024llama}, Qwen 2 series \citep{yang2024qwen2}, and GLM series \citep{glm2024chatglm}, we find that their average performance lags behind human performance by 44.7\%, with the gap widening significantly when requiring models to track how a mental state changes across different scenarios. This performance degradation highlights a fundamental limitation in current LLMs' ability to model the dynamic nature of human mental states.

The main contributions of this work are: 1. A novel framework for evaluating LLMs' understanding of temporal evolution in mental states, with a systematic process for generating and validating evaluation data; 2. A comprehensive benchmark featuring 78,100 questions specifically designed to probe LLMs' ability to track and reason about mental state changes over scenarios; 3. Extensive empirical evaluation reveals specific challenges LLMs face in temporal reasoning about mental states, including detailed analysis of failure modes in tracking state changes and determining factors influencing changes.

%% file: section/relatedwork.tex
\section{Related Work}
\subsection{ToM Benchmarks}
Theory of mind appears to be an innate potential ability in humans that requires social and other experiences over many years for its full development. With the development of LLMs, researchers have begun to probe whether LLMs possess a Theory of Mind ability comparable to that of humans, as they have reached and occasionally surpassed human performance in some task-solving and reasoning tasks. \citet{nematzadeh-etal-2018-evaluating,le-etal-2019-revisiting,wu-etal-2023-hi} apply the Sally-Anne Test and bAbi to test LLMs' ToM ability in the aspect of false belief, and they find that LMs' performance is significantly lower than humans. \citet{ullman2023large,shapira-etal-2024-clever,kim-etal-2023-fantom,sap-etal-2022-neural} propose that LLMs prone to shortcuts and spurious correlations. Apart from the test in the aspect of belief, \citet{xu-etal-2024-opentom,chen-etal-2024-tombench,sabour-etal-2024-emobench} construct benchmarks to test LLMs' ToM ability for emotion, intention, and perception. \citet{jin-etal-2024-mmtom,shi2024mumatommultimodalmultiagenttheory} propose to evaluate LLMs in multi-modal environments. However, most of the previous evaluations do not take the continuous evolution of mental states across multiple interconnected scenarios into consideration. Our work aims to develop a novel benchmark to understand the ToM reasoning of language models in the dynamic social context.

\subsection{Human Behavior Simulation}

Recent advancements in language model capabilities have opened new avenues for generating high-quality data. Previous work has demonstrated successful applications of LLMs in simulating human behavior across various domains, including HCI research \citep{10.1145/3544548.3580688}, conversational recommender systems \citep{yoon-etal-2024-evaluating}, role-playing \citep{xie2024can,xiao2023far}, clinical medical education \citep{wang2024towards}, social science \citep{hua2023war,park2023generative,park2022social,10.5555/3618408.3618425}. \textsc{DynToM} leverages LLMs to generate realistic dialogues that reflect predetermined character mental states. We implement strict quality assurance through human evaluation of social context authenticity and question validity. This approach combines the efficiency of automated generation with robust validation procedures, ensuring our benchmark's reliability and reproducibility.

%% file: section/evaluation2construction.tex
\section{\textsc{DynToM} Benchmark}
\subsection{\textsc{DynToM} Construction Framework}
\label{sec:DynToM}
% \paragraph{Definitions and Preliminaries} 
% We first define key terms used throughout this paper. A \textit{Social Location} refers to the physical setting where social interactions occur, which influences behavior and social norms. A \textit{Social Scenario} represents a single, specific social interaction at a given moment, unlike previous works where scenarios might span multiple events. A \textit{Social Context} encompasses both the social background (location and character profiles) and a series of connected social scenarios.

% \paragraph{Definitions and Preliminaries} 
% We first define key terms used throughout this paper. A \textit{Social Location} refers to the physical setting where social interactions occur, which influences behavior and social norms \cite{farrow2017social}. A \textit{Social Context} encompasses a social background, which includes both the social location and characters and their profiles (e.g., their demographics, personalities, and relationships). A \textit{Social Scenario} represents a specific social interaction occurring within a social context at a given moment. Multiple social scenarios, happening at different times within the same social context, capture the dynamic evolution of characters' mental states as they interact over time. This temporal sequence of scenarios allows us to track how beliefs, emotions, and intentions change through ongoing social interactions.

\paragraph{Definitions and Preliminaries} 
We first define key terms used throughout this paper. A \textit{Social Location} refers to the physical setting where social interactions occur, which influences behavior and social norms \citep{farrow2017social}. A \textit{Social Context} provides the foundational setup for social interactions, comprising a social location, character profiles (e.g., demographics, personalities), and their relationships. A \textit{Social Scenario} represents a self-contained social interaction between characters at a specific moment. In our work, we construct sequences of temporally connected scenarios within the same social context, enabling us to track the dynamic evolution of characters' mental states through continuous social interactions. We define a \textit{Social Stage} as the complete structure of a social interaction, comprising the \textit{Social Location}, \textit{Social Context}, and \textit{Social Scenarios}.

Our framework consists of three systematic steps for generating the social stages in our benchmark:

\paragraph{Step 1: Social Context Construction} 
A social context consists of three components: a social location, two characters' profiles, and the relationship between these characters. For social locations, following \citet{ziems-etal-2023-normbank}, we collect 261 locations across 13 categories representing common physical settings for social interactions. For character profiles, we construct seven aspect pools (names, surnames, gender, occupation, education, race, and personality traits) using demographic data from the U.S. Census Bureau statistics to ensure realistic population representation. For each social context, we randomly sample one location and create two characters by sampling from each aspect of these pools. To generate character relationships, we first create four exemplar relationships manually, then prompt GPT-4-Turbo to generate new relationships based on these exemplars and the sampled character profiles. To ensure quality, four human annotators evaluate both the characters' profiles and their corresponding relationships, discarding any profile or relationship that any annotator deems unrealistic. This rigorous validation process results in retaining 92\% of the generated profiles and relationships.

\paragraph{Step 2: Mental State Trajectory Design}
We focus on evaluating three mental states (\textbf{beliefs}, \textbf{emotions}, and \textbf{intentions}) and their resulting \textbf{actions} (for convenience, also denoted as mental states). For each social context, we design a sequence of five\footnote{we have limited the number of scenarios to five to reduce costs while maintaining a more authentic social context. Researchers can easily adjust the scenario number in our framework to meet their needs.} scenarios where these states of the characters evolve and influence each other. Following the psychological research of \citet{d1995development}, we model the mental states through three key relationships: 1) beliefs influence emotions; 2) beliefs and emotions influence intentions; 3) beliefs, emotions, and intentions drive actions. We prompt GPT-4-Turbo with four exemplar trajectories and the 3 design principles to generate coherent mental state progressions across every five scenarios. Importantly, when generating these trajectories, LLMs should also output specific cues that trigger mental state transitions between adjacent scenarios, providing explicit reasoning for how and why mental states evolve throughout the social interaction. Any scenarios without the mental state trajectory and cues are discarded. Four human annotators evaluate each generated trajectory on two dimensions using a 5-point scale: coherence (consistency of mental state changes across scenarios), rationality (the validity and rationality of these transition cues), and authenticity (plausibility of mental state transitions). Trajectories with mean scores below 4.0 on either dimension are discarded, resulting in an 85.4\% retention rate.

\paragraph{Step 3: Scenario Generation}
Building upon the mental state trajectories designed in Step 2, we now generate scenarios with natural dialogues to manifest these mental states in social interactions. Each scenario includes a background description and a dialogue between characters, reflecting the mental state trajectory designed for this scenario. We choose dialogue as the primary format because it naturally reveals characters' mental states and is frequently used in daily interactions. For each mental state trajectory, we prompt GPT-4-Turbo to generate the dialogue and background of the scenario, ensuring that the main character's utterances and behaviors align with their prescribed mental states.

Following the same validation process as Step 2, four human annotators evaluate each scenario on three dimensions using a 5-point scale: consistency (alignment with the designed mental state trajectory), coherent (the five scenarios within each social context form a coherent storyline, where each scenario connects meaningfully to those before and after it.), and authenticity (naturalness of the scenario and conversations). Scenarios scoring below 4.0 on either dimension are discarded, with 88.7\% of the generated scenarios retained.

\subsection{Question Genres}

\begin{figure}[t]
    \centering
    \includegraphics[width=\hsize]{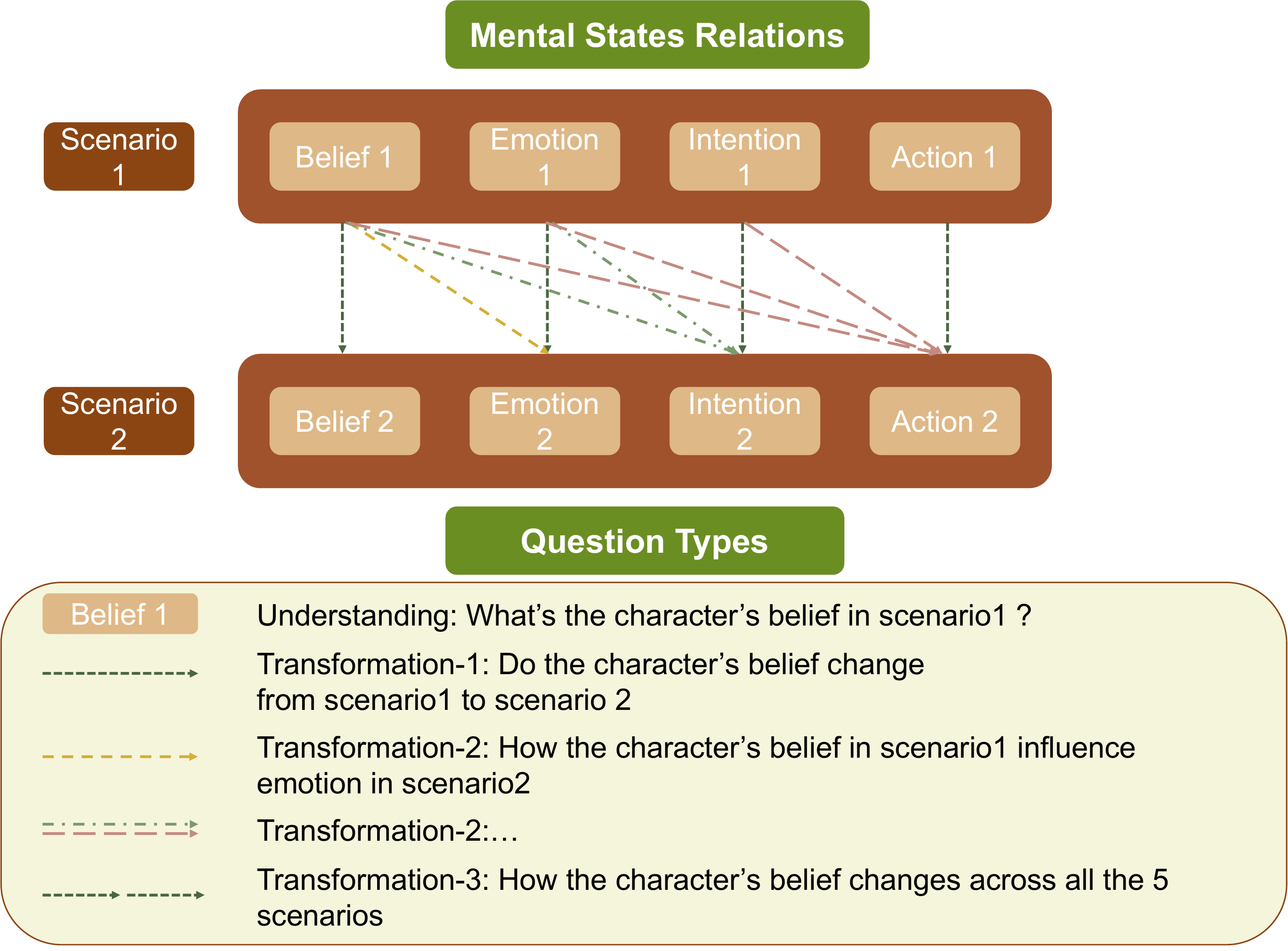}
    \caption{A simplified example of question types (showing only two scenarios). In each scenario, characters have four states: belief, emotion, intention, and action - these are assessed through \textbf{understanding} questions. The value of the same state changes across different times (scenarios) is evaluated through \textbf{transformation-1,2,3} questions.}
    \vspace{-10pt} 
    \label{fig: question_types}
\end{figure}

% \paragraph{Step 4: Question Generation}

Based on the validated scenarios and mental state trajectories, we design questions to systematically evaluate how well LLMs can track and reason about the temporal evolution of mental states. We develop four question types that progressively assess different aspects of this capability.

\paragraph{Understanding Questions}
Understanding questions establish a baseline by testing LLMs' ability to identify states (belief, emotion, intention, and action) at specific points in time. While this represents the most basic level of ToM reasoning in our evaluation, it forms the foundation for evaluating more complex temporal understanding.
\paragraph{Transformation Questions}
Transformation questions directly evaluate LLMs' ability to reason about mental state dynamics through three increasingly complex aspects, as illustrated in Figure \ref{fig: question_types}. \textbf{Transformation-1} examines whether LLMs can detect state changes between consecutive scenarios, testing their basic temporal awareness. \textbf{Transformation-2} probes deeper by testing if LLMs understand the causal mechanisms behind state changes, evaluating their grasp of psychological dynamics. \textbf{Transformation-3} presents the most challenging task: tracking state evolution across all scenarios and testing LLMs' ability to maintain and reason about extended temporal sequences.

This progression of question types allows us to precisely identify where LLMs succeed or fail in understanding dynamic mental states. Understanding questions reveal whether failures in temporal reasoning stem from basic state comprehension issues, while the three transformation types help pinpoint specific limitations in tracking and reasoning about mental state changes over time. We apply four predefined question templates to the social stage to generate questions. The template details are presented in the Appendix \ref{append:question template and examples}. 

\paragraph{Options and Ground Truth}
The design of options and ground truth leverages the comprehensive mental state trajectories created in Step 2. For understanding questions, when evaluating a specific state (e.g., belief in scenario 1), we construct distractors using both other states from the same scenario and the same state from other scenarios. For instance, when assessing belief in scenario 1, incorrect options include the emotion, intention, and action from scenario 1, as well as beliefs from scenarios 2-5. Similarly, for transformation questions, options are constructed using documented state values and their changes from the trajectory. For example, when asking "Why does John's belief change from feeling inferior in scenario 1 to feeling respected in scenario 2?", the correct answer would be "Beverly's praise of his expertise," while distractors include other documented changes such as "John's demonstration of skills" (action). This systematic approach of option generation ensures that the questions are challenging yet unambiguous, as both correct answers and distractors are grounded in the explicitly designed mental state trajectories.

\paragraph{Validation}
Following our validation process, four annotators evaluate the questions on clarity (whether the question is unambiguous) and answerability (whether the answer can be determined from the given context) using a 5-point scale. Questions scoring below 4.0 are regenerated, resulting in a final set of high-quality evaluation items. Finally, 78100 questions are collected.

\paragraph{Evaluation Metric}
To evaluate LLMs' performance on \textsc{DynToM}, we calculate their accuracy across all 78,100 questions. For each question, we consider the LLM's response correct only if it exactly matches the ground truth option. The final performance metric is computed as the percentage of correct answers across all questions, providing a comprehensive measure of the model's ability to reason about dynamic mental states in social interactions.

\subsection{Statistics}

In total, our final benchmark contains 1,100 high-quality social stages, where each stage consists of a social location (physical setting), social context (two characters with detailed profiles and relationships), and five social scenarios (temporally connected interactions). For each social stage, we generate 71 questions across four types (understanding and three types of transformation), resulting in 78,100 questions in total. The detailed statistics are shown in Table \ref{tab:tomvalley-stats}.

Compared to existing ToM benchmarks (Table \ref{tab:benchmark_comparison}), \textsc{DynToM} offers key advantages. While previous works evaluate static snapshots of mental states, \textsc{DynToM} systematically captures the temporal evolution of mental states through connected scenarios, enabling a more realistic evaluation of how well LLMs can track and reason about dynamic social interactions.

\begin{table}[t]
\centering
\small
\begin{tabular}{lr}
\toprule
Item & Number \\
\midrule
ToM Mental States & 4 \\
Social Locations & 261 \\
Characters & 2200 \\
Social Scenarios & 5500 \\
Social Contexts & 1100 \\
\midrule
Questions & 78100 \\
\quad Understanding & 28.2\% \\
\quad Transformation-1 & 22.5\% \\
\quad Transformation-2 & 43.7\%\\
\quad Transformation-3 & 5.6\% \\

\midrule
Average Social Scenario Length & 457.9 \\
Average Questions Length & 77.5 \\
\bottomrule
\end{tabular}
\caption{\textsc{DynToM} Statistics.}
\vspace{-10pt} 
\label{tab:tomvalley-stats}
\end{table}

\begin{table}[t]
\small
    \adjustbox{width=\hsize}{
    \centering
        \begin{tabular}{l|cccccc|c}
                \toprule
                \multicolumn{2}{l}{{\color{1}  \faHome} Social Location} &
                \multicolumn{3}{l}{{\color{3}  \faUserFriends} Relationship} &
                \multicolumn{3}{l}{{\color{5} \faLink} Dynamic mental states}
                \\
                %\multicolumn{4}{l}{{\color{3}  \faUserFriends} : Relationship} &
                \multicolumn{1}{l}{{\color{2}  \faIdCard}  Profile} & 
                \multicolumn{5}{l}{{\color{4}  \faStaylinked}  Intradependent mental states}
                & \multicolumn{2}{l}{{\color{6} \faPenNib}  Questions Num} \\
                %\multicolumn{4}{l}{{\color{5} \faLink} : \makecell[l]{Dy-
                %namic\\ mental states}} &
                %\multicolumn{4}{l}{{\color{6} \faPenNib} : \makecell[l]{Number of Questions}} \\
                \toprule
                \toprule
                & Plot & 
                {\color{1} \faHome} & 
                {\color{2} \faIdCard} & 
                {\color{3} \faUserFriends} & 
                {\color{4} \faStaylinked} & 
                {\color{5} \faLink} & 
                {\color{6} \faPenNib} 
                \\
                \midrule
                ToMi & \ding{56} & \ding{56} & \ding{56} & \ding{56} & \ding{56} & \ding{56} & 999 \\
                SocialIQA & \ding{56} & \ding{56} & \ding{56} & \ding{56} & \ding{56} & \ding{56} & 37588 \\
                Hi-ToM & \ding{56} & \ding{52} & \ding{56} & \ding{56} & \ding{56} & \ding{56} & 1200 \\
                OpenToM & \ding{52} & \ding{56} & \ding{52} & \ding{52} & \ding{56} & \ding{56} & 2384 \\
                BigToM & \ding{52} & \ding{52} & \ding{56} & \ding{56} & \ding{56} & \ding{52} & 600 \\
                TOMBENCH & \ding{52} & \ding{56} & \ding{56} & \ding{56} & \ding{56} & \ding{56} & 2860 \\
                \midrule 
                \midrule 
                \textsc{DynToM}(ours) & \ding{52} & \ding{52} & \ding{52} & \ding{52} & \ding{52} & \ding{52} & 78100 \\
                \bottomrule
            \end{tabular}}
    
    \captionof{table}{ Benchmark Comparison.}
    \vspace{-10pt} 
    \label{tab:benchmark_comparison}
    % \vspace{-10pt}
\end{table}

%% file: section/experiments.tex
\section{Experiments}
\subsection{Experimental Setup}
To evaluate the ToM reasoning capabilities across different model scales and architectures, we conducted experiments using \textsc{DynToM} on ten representative language models, ranging from 7B to 70B parameters. These models include GPT-4o, GPT-4-Turbo, Llama-3.1 (8B and 70B variants), Mistral-7B, Mixtral-8x7B, Qwen2 (7B and 72B variants), DeepSeek-V2, and GLM-4. All models are accessed through their official APIs or publicly available weights.

We employed two evaluation approaches: (1) vanilla prompting, where models directly answer questions, and (2) zero-shot chain-of-thought (CoT) prompting \citep{wei2022chain}, which encourages step-by-step reasoning before providing final answers. For both vanilla and CoT prompting, we used a temperature of 0.7 and top-p of 0.9 across all models to ensure fair comparison. To establish a human performance baseline, we recruited ten graduate students, different from those involved in data annotation, to evaluate a randomly sampled 30\% of the dataset (330 social stages and 23430 questions). Detailed specifications of model versions, architectures, context windows, and prompting templates are provided in Appendix \ref{append:model detail}.

\subsection{Main Results}
Table \ref{tab:model_performance} demonstrates the ToM performance of LLMs across different mental states (belief, emotion, intention, and action) and question types (understanding and transformation), both with and without chain-of-thought (CoT) prompting. We established the human baseline by averaging performance across ten annotators, with standard deviations reported to indicate inter-annotator agreement. Here, we discuss several key findings from our experimental results.

\begin{table*}[t]
\adjustbox{width=\textwidth}{
\centering
\small
\begin{tabular}{lccccccccl}
\toprule
\multirow{2}{*}{Subject} & \multicolumn{2}{c}{Belief} & \multicolumn{2}{c}{Emotion} & \multicolumn{2}{c}{Intention} & \multicolumn{2}{c}{Action} & \multirow{2}{*}{AVG.} \\
\cmidrule(lr){2-3} \cmidrule(lr){4-5} \cmidrule(lr){6-7} \cmidrule(lr){8-9}
& U & T & U & T & U & T & U & T & \\
\midrule
Human & 83.8$_{\scriptstyle\pm16.4}$ & 77.6$_{\scriptstyle\pm12.0}$ & 89.5$_{\scriptstyle\pm10.7}$ & 78.7$_{\scriptstyle\pm14.0}$ & 79.0$_{\scriptstyle\pm21.4}$ & 73.8$_{\scriptstyle\pm14.0}$ & 76.7$_{\scriptstyle\pm25.8}$ & 76.3$_{\scriptstyle\pm14.0}$ & 77.7$_{\scriptstyle\pm12.7}$ \\
\midrule
GPT-4o & \textbf{80.9} & \textbf{44.5} & 91.7 & \textbf{45.8} & \textbf{87.5} & \textbf{51.9} & \textbf{95.1} & \textbf{55.6} & \textbf{64.0} \\
GPT-4-Turbo & 63.5 & 32.3 & 74.7 & 33.9 & 71.3 & 35.5 & 80.5 & 36.2 & 47.6 \\
Llama-3.1-70B & 65.8 & 40.2 & \textbf{93.8} & 42.3 & 82.8 & 42.0 & 91.8 & 45.5 & 57.1 \\
Llama-3.1-8B & 31.6 & 18.0 & 40.0 & 19.9 & 22.4 & 16.6 & 26.6 & 15.5 & 22.3 \\
Mixtral-8x7B & 23.3 & 21.6 & 46.2 & 18.4 & 32.9 & 10.8 & 40.3 & 9.5 & 21.9 \\
Mistral-7B & 21.3 & 11.7 & 23.8 & 10.2 & 16.3 & 10.1 & 20.6 & 9.2 & 13.9 \\
Qwen2-72B & 72.0 & 37.2 & 85.5 & 38.0 & 79.5 & 33.2 & 89.8 & 20.9 & 48.5 \\
Qwen2-7B & 22.2 & 19.8 & 43.0 & 20.5 & 25.1 & 15.7 & 24.6 & 15.0 & 22.1 \\
DeepSeek-V2 & 6.5 & 9.2 & 4.8 & 8.1 & 3.7 & 7.3 & 2.8 & 5.7 & 7.2 \\
GLM-4 & 29.5 & 23.9 & 43.9 & 20.8 & 28.5 & 16.5 & 40.4 & 16.8 & 25.4 \\
\midrule
LLM AVG. & 41.7 & 25.8 & 54.7 & 25.8 & 45.0 & 24.0 & 51.3 & 23.0 & 33.0 \\
\midrule
GPT-4o+CoT & \textbf{79.2} & \textbf{44.5} & 88.0 & \textbf{47.6} & 82.1 & \textbf{46.6} & 90.4 & \textbf{49.6} & \textbf{61.1} \\
GPT-4-Turbo+CoT & 61.7 & 31.0 & 77.8 & 33.2 & 71.4 & 32.8 & 81.0 & 37.6 & 47.1 \\
Llama-3.1-70B+CoT & 68.0 & 38.9 & \textbf{90.7} & 43.7 & 81.4 & 42.8 & \textbf{96.5} & 46.6 & 57.6 \\
Llama-3.1-8B+CoT & 32.0 & 21.7 & 40.3 & 20.9 & 21.8 & 19.3 & 23.3 & 15.9 & 23.6 \\
Mixtral-8x7B+CoT & 15.6 & 13.9 & 29.7 & 13.8 & 25.8 & 8.8 & 26.6 & 8.8 & 15.8 \\
Mistral-7B+CoT & 21.6 & 10.1 & 22.5 & 11.0 & 19.9 & 8.1 & 18.8 & 8.8 & 13.3 \\
Qwen2-72B+CoT & 70.1 & 39.2 & 87.6 & 41.4 & \textbf{83.8} & 34.6 & 89.0 & 27.1 & 51.3 \\
Qwen2-7B+CoT & 28.6 & 18.1 & 43.7 & 19.3 & 29.6 & 19.7 & 20.2 & 18.4 & 23.5 \\
DeepSeek-V2+CoT & 7.4 & 9.8 & 3.2 & 10.4 & 5.0 & 7.3 & 5.0 & 6.4 & 8.1 \\
GLM-4+CoT & 30.0 & 26.4 & 48.0 & 22.1 & 32.4 & 17.7 & 43.2 & 14.1 & 26.6 \\
\midrule
LLM+CoT AVG. & 41.4 & 25.4 & 53.2 & 26.3 & 45.3 & 23.8 & 49.4 & 23.3 & 32.8 \\
\bottomrule
\end{tabular}}
\caption{LLMs' performance on \textsc{DynToM}. U: Understanding, T: Transformation. Numbers represent accuracy in percentages. For human performance, subscripts indicate standard deviation across ten annotators.}
\vspace{-10pt} 
\label{tab:model_performance}
\end{table*}

% \paragraph{Human vs. LLMs}
% Human annotators achieved an average accuracy of 77.7\%$_{\scriptstyle\pm12.7}$ across all tasks. All LLMs performed significantly below this baseline, with GPT-4o showing the smallest gap at 13.7\% in vanilla prompting (77.7\% vs. 64.0\%). The performance disparity was particularly pronounced in transformation-type questions across all mental states, suggesting that dynamic reasoning about mental state changes remains a significant challenge for current LLMs. Notably, however, some LLMs, particularly GPT-4o, demonstrated superior performance in understanding-type questions compared to human annotators. This can be attributed to the nature of these questions, which primarily involve direct extraction of explicitly stated mental states within single scenarios, as illustrated in Figure \ref{fig: pipeline}. Such tasks rely more on semantic matching than complex reasoning, aligning well with LLMs' strong pattern recognition capabilities.

\paragraph{Human vs. LLMs}
Human annotators achieved an average accuracy of 77.7\% across all tasks. All LLMs performed significantly below this baseline, and their average performance underperforms humans by 44.7\%, with even the best GPT-4o showing a gap up to 13.7\% in vanilla prompting (77.7\% vs. 64.0\%). The performance disparity was particularly pronounced in transformation-type questions across all mental states, revealing current LLMs' limitations in tracking and comprehending the temporal evolution of mental states in social interactions. Notably, however, some LLMs, particularly GPT-4o, demonstrated superior performance in understanding-type questions compared to human annotators. This can be attributed to the nature of these questions, which primarily involve analyzing static mental states within single scenarios, as illustrated in Figure \ref{fig: question_types}. Such tasks require less temporal reasoning and rely more on semantic matching of explicitly stated mental states, aligning well with LLMs' pattern recognition capabilities but not necessarily reflecting the true understanding of dynamic social contexts.

\paragraph{Differences Between LLMs' ToM Performance}
In vanilla prompting settings, GPT-4o emerged as the leading model, achieving an accuracy of 64.0\% and outperforming the second-best model, Llama-3.1-70B (57.1\%), by 6.9 percentage points. Among open-source models, both Llama-3.1-70B and Qwen2-72B demonstrated remarkable capabilities, surpassing GPT-4-Turbo's 47.6\% performance. Notably, Llama-3.1-70B achieved superior performance in emotion-related understanding tasks, reaching 93.8\% accuracy compared to GPT-4o's 91.7\%. However, even GPT-4o's best overall performance at 64.0\% falls significantly short of human-level performance, with DeepSeek-V2 showing the lowest performance at 7.2\%. This substantial performance gap, particularly in tracking the temporal evolution of mental states, highlights the challenging nature of our benchmark and reveals that current LLMs lack robust ToM reasoning capabilities in realistic social contexts, despite their near-perfect performance on existing ToM benchmarks \citep{gandhi2024understanding}.

\paragraph{Differences Between Transformation and Understanding Question Types}  
Table \ref{tab:model_performance} also reveals that models perform significantly worse on transformation questions compared to understanding questions. The most substantial gap occurs in emotion-related reasoning, where the average model accuracy drops from 54.7\% in understanding questions to 25.8\% in transformation questions—a difference of 28.9 percentage points. Transformation questions require models to track how a character’s mental state evolves across different scenarios, capturing shifts in beliefs, emotions, and intentions over time. This performance gap highlights a critical limitation of current models—their inability to effectively reason about dynamic mental state changes within continuous social contexts.

\paragraph{Vanilla vs. CoT Prompting}
Our experimental results demonstrate that standard chain-of-thought (CoT) prompting has inconsistent effects on LLMs' ToM reasoning capabilities. While CoT prompting improved performance for smaller models (Llama-3.1-8B: +1.3\%, Qwen2-72B: +2.8\%, DeepSeek-V2: +0.9\%, and GLM-4: +1.2\%), it led to performance degradation in more capable models, notably GPT-4o (-2.9\%). Through analyzing the intermediate outputs (Appendix \ref{append:CoT_case_study}), we identified a two-fold effect: For smaller models that struggle with complex reasoning, CoT's step-by-step decomposition provides beneficial scaffolding for basic problem analysis. However, this same decomposition becomes a limitation for more capable models, as it enforces a rigid reasoning structure that treats each scenario independently, failing to capture the crucial temporal dependencies between scenarios. As shown in our case study, when asked about mental state changes from scenarios 1 to 2, models following CoT often extensively analyze each scenario but fail to explicitly compare the states across time steps, leading to incorrect conclusions. This aligns with findings in \citet{xiao2023far} about LLMs' challenges in maintaining coherence during reasoning long inputs. These observations suggest that while CoT can help with basic reasoning decomposition, effective ToM reasoning requires specialized promptings that explicitly guide models to track and analyze the temporal evolution of mental states across scenarios.

\paragraph{Differences Across Mental States}
Analysis of Table \ref{tab:model_performance} reveals consistent patterns in models' capability to reason about different mental states, particularly in understanding-type questions. Emotion-related reasoning achieves the highest accuracy, averaging 54.7\%, whereas belief-related reasoning lags behind at 41.7\%—a gap of 13 percentage points. We hypothesize that this disparity stems from the inherently implicit nature of beliefs compared to other mental states – while emotions and intentions often manifest in observable behaviors or explicit statements, beliefs frequently require multi-step inference from indirect evidence, such as actions or conversational context. This observation suggests that belief reasoning poses unique challenges in temporal social contexts, where models must not only infer current beliefs but also track their evolution through sequential interactions.

\subsection{In-Depth Analysis}

\paragraph{LLMs' Limits of ToM on Transformation}

\begin{figure}[t]
    \centering
    \small
    \includegraphics[width=\hsize]{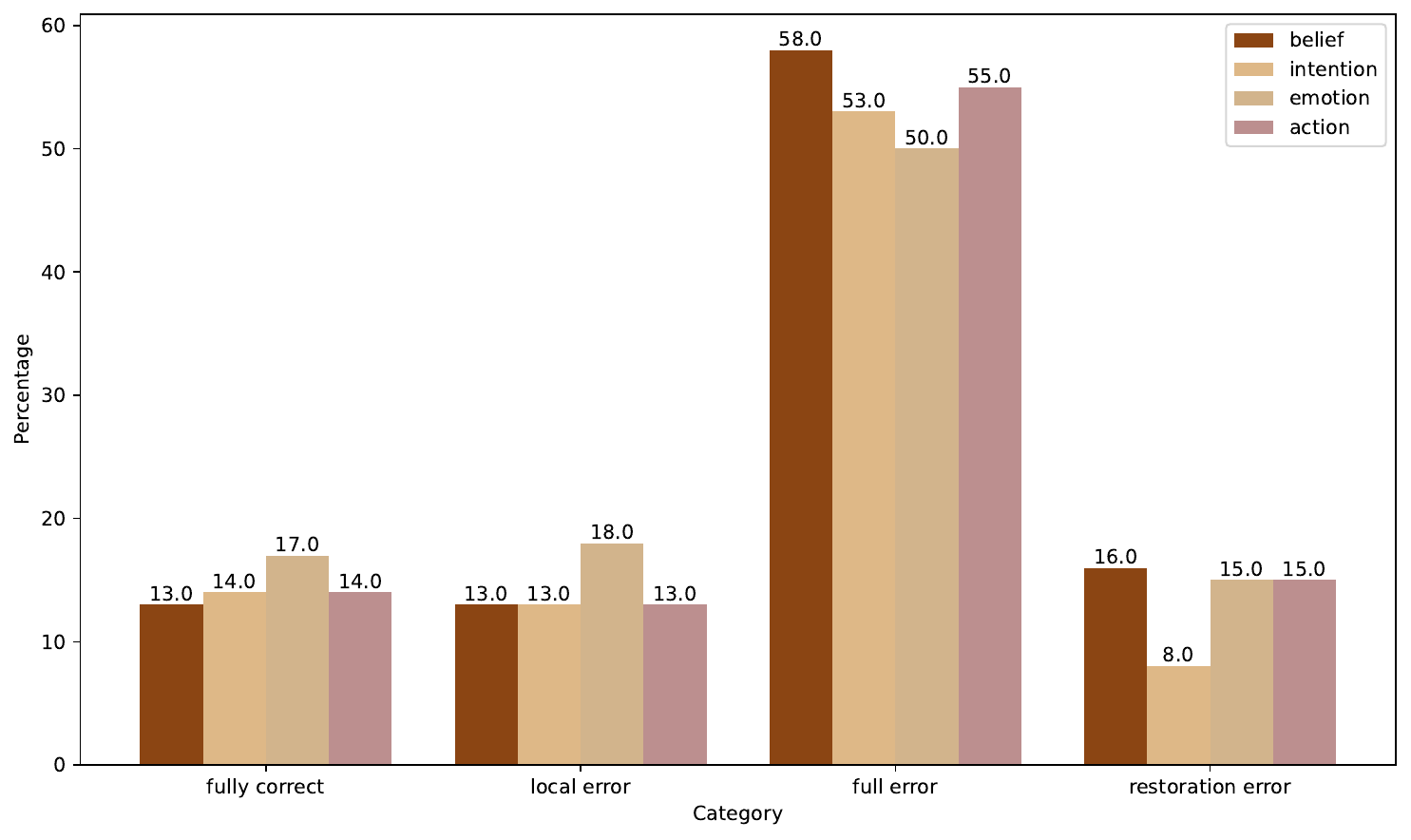} % Replace with the correct path
    \caption{The percentage of GPT-4o's four types of their response. }
    \vspace{-10pt}
    \label{fig:complex}
\end{figure}

Evaluating mental states across multiple interconnected scenarios introduces complex compositional reasoning challenges \citep{dziri2023faith}, requiring models to track and reason about the continuous evolution of mental states. To systematically analyze how models handle this multi-step reasoning process, we group related questions and categorize the model's responses based on their performance on both the primary question and its dependencies.
For instance, consider a question that asks whether a character's beliefs change between two scenarios. To answer this primary question (denoted as C), the model must first correctly identify the character's beliefs in each individual scenario (denoted as D). By grouping such related questions, we can assess the model's responses and classify them into four types: (1) Fully correct: The model accurately answers both the primary question (C) and all its dependencies (D). (2) Local error: The model correctly answers all dependencies (D) but makes an error on the primary question (C). (3) Restoration error: The model correctly answers the primary question (C) despite making errors on one or more of its dependencies (D). (4) Full error: The model makes errors on both the primary question (C) and one or more of its dependencies (D).

Calculating the proportion of each response type across all question groups provides a comprehensive evaluation of the model's performance on these compositional questions requiring multi-step reasoning. As shown in Figure \ref{fig:complex}, our analysis reveals distinct patterns in models' reasoning capabilities. The fully correct cases, where models successfully identify both the mental states and their changes, are notably rare across all state types (13-17\%), indicating limited genuine understanding. Local errors (13-18\%) show models can correctly identify mental states but fail to reason about their changes, suggesting an inability to track evolution. Full errors dominate across all mental states (50-58\%), with belief states showing the highest error rate (58\%), revealing fundamental limitations in comprehending both states and their transitions. Restoration errors (8-16\%) occur when models correctly identify changes without understanding the underlying states, indicating superficial pattern matching rather than genuine reasoning. This analysis pinpoints where models struggle in the reasoning process, whether in identifying initial mental states, determining factors influencing changes, or integrating information across scenarios.

\begin{figure}[t]
    \centering
    \small
    \includegraphics[width=\hsize]{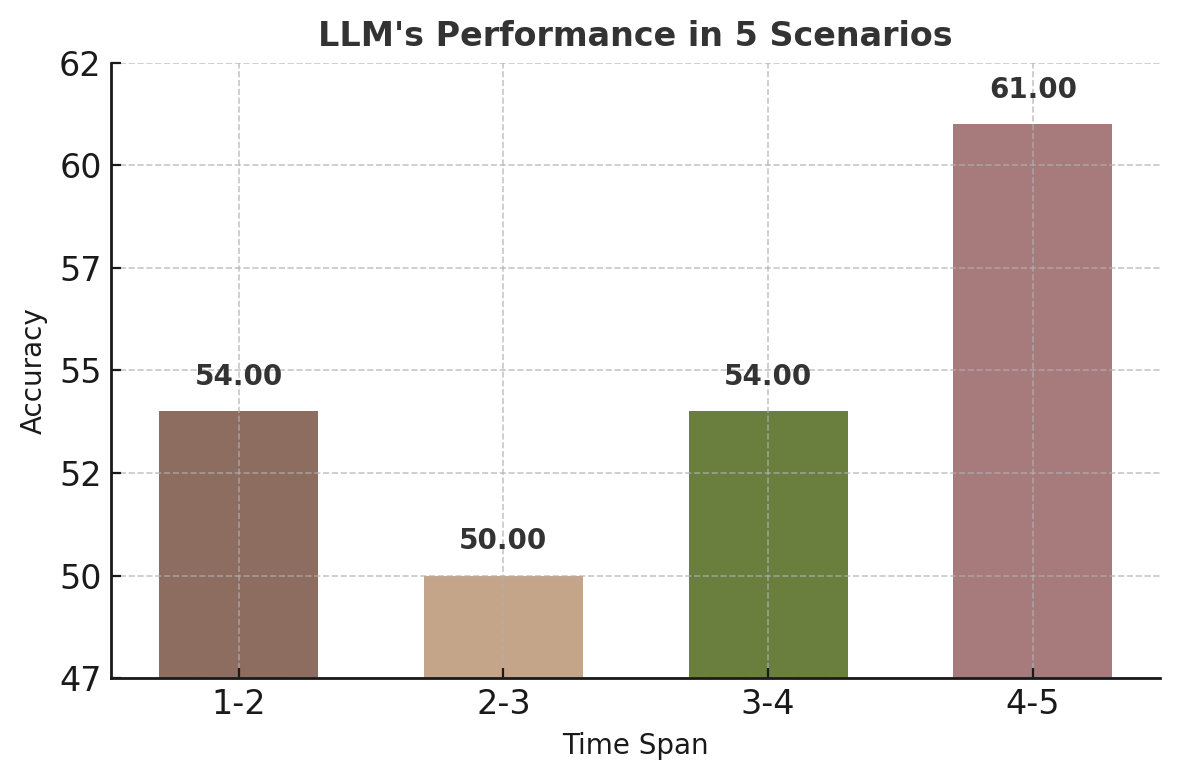} % Replace with the correct path
    \caption{The average of GPT-4o's scores of the transformation question type in different time spans. The time span indicates the specific scenarios to which one question relates.}
    \vspace{-10pt} 
    \label{fig:time_span_accuracy}
\end{figure}

\paragraph{LLMs Fail in the Middle Scenario}
\label{sec:LLM_fail_middle}

Our analysis reveals a consistent pattern of lower performance in transformation-type questions compared to understanding-type questions across all models. To investigate this performance gap, we examine model accuracy across different time spans, where each span represents the interval between consecutive scenarios (e.g., span 1-2 represents the transition from scenarios 1 to 2). We categorize the questions into different time spans based on their associated social scenarios. The results show a distinct "U-shaped" pattern: models perform better at early and late time spans but struggle with middle spans.
To validate whether this pattern stems from the "Lost in the middle" phenomenon \citep{liu-etal-2024-lost} - where model performance degrades when processing information from the middle of long contexts - we conducted two additional experiments. First, we extend our analysis to longer sequences of 6 and 7 scenarios. The results (Table \ref{table:time span}) strongly support our hypothesis: in 6-scenario sequences, accuracy drops to 50\% in span 2-3, while in 7-scenario sequences, performance deteriorates more severely to 26\% in span 3-4. Second, we perform a controlled experiment where we truncate sequences to their first four scenarios. As shown in Table \ref{table:time span alation}, this intervention leads to significant performance improvements in middle spans: for 7-scenario sequences, the accuracy in span 3-4 improve by 21 percentage points (from 26\% to 47\%) when later scenarios are removed. Similar improvements are observed across 5- and 6-scenario sequences, with middle-span accuracy increasing by 1-5 percentage points. 

These consistent improvements across different sequence lengths reveal a critical limitation in LLMs' ability to process dynamic mental states. While models can effectively track states at the beginning and end of interactions, they struggle to maintain understanding across extended scenarios - precisely the kind of continuous evolution that characterizes real-world social interactions. The significant performance degradation in middle scenarios (dropping to as low as 26\% in longer sequences) underscores the importance of our dynamic evaluation approach and highlights a fundamental challenge in developing LLMs that can truly understand evolving social contexts.

\begin{table}[t]
\centering
\small
\begin{tabular}{lcccccc}
\toprule
Time Span & 1-2 & 2-3 & 3-4 & 4-5 & 5-6 & 6-7 \\
\midrule
6 Scenarios & 64.0 & \textbf{50.0} & 51.0 & 62.0 & 62.0 & - \\ 
7 Scenarios & 56.0 & 45.0 & \textbf{26.0} & 30.0 & 26.0 & 34.0 \\ 
\bottomrule
\end{tabular}
\caption{The average of GPT-4o's scores in the transformation question type for 6 and 7 scenarios.\label{table:time span}}
\vspace{-10pt}
\end{table}

\begin{table}[t]
\adjustbox{width=0.5\textwidth}{
\centering
\small
\begin{tabular}{lcccc}
\toprule
Category & Time Span & \makecell{w/o the \\truncation} & \makecell{w/ the\\ truncation} & \textbf{$\Delta$} \\
\midrule
\multirow{3}{*}{5 scenarios} & \textbf{1-2} & 54.0 & 55.0 & +1.0 \\ 
 & \textbf{2-3} & 50.0 & 53.0 & +3.0 \\ 
 & \textbf{3-4} & 54.0 & 55.0 & +1.0 \\ 
 \midrule
\multirow{3}{*}{6 scenarios} & \textbf{1-2} & 60.0 & 60.0 & +0.0 \\ 
 & \textbf{2-3} & 50.0 & 54.0 & +4.0 \\
 & \textbf{3-4} & 51.0 & 56.0 & +5.0 \\ 
 \midrule
\multirow{3}{*}{7 scenarios} & \textbf{1-2} & 56.0 & 53.0 & -3.0 \\ 
 & \textbf{2-3} & 45.0 & 54.0 & +9.0 \\ 
 & \textbf{3-4} & 26.0 & 47.0 & +21.0 \\ 
 \bottomrule
\end{tabular}}
\caption{Comparison of GPT-4o's performance with and without the truncation of the fifth/seventh/eighth scenario across different time spans and categories (total 5, 6, and 7 scenarios), with Delta indicating the difference between the two conditions.\label{table:time span alation}}
\vspace{-10pt}
\end{table}

\section{Conclusion}
In this work, we present DynToM, a benchmark designed to evaluate LLMs' theory of mind capabilities in dynamic social contexts, moving beyond static assessments to capture the crucial evolution of mental states across interconnected scenarios. Our human evaluation validates the benchmark's alignment with real-world social dynamics, while experimental results reveal significant gaps in current LLMs' capabilities - even the best model trails human performance by 11\%, particularly struggling with tracking mental state changes across extended interactions. As LLMs continue to be deployed in social contexts, DynToM provides a valuable framework for assessing and improving their ability to understand the dynamic nature of human mental states.

%% file: section/discussion.tex
\section*{Limitations}
\paragraph{Limited LLMs} This paper makes a significant contribution to the field by introducing \textsc{DynToM}, an innovative benchmark designed to evaluate Theory of Mind capabilities in language models within authentic social contexts. However, due to the constraint of computing resources and budget, a limitation lies in its evaluation scope, which encompasses ten language models with an emphasis on representative models. While this selection includes prominent models such as GPT-4 and Llama, the focus potentially overlooks insights that could be gained from examining other emerging open-source models and commercial models, such as Claude. 

\paragraph{Limited Prompt Methods}
We use vanilla and CoT prompting methods for evaluation. Other methods, such as think-twice \citep{wilf-etal-2024-think}, belief tracker \citep{sclar-etal-2023-minding}, and self-consistency \citep{wang2023selfconsistency}, could also be explored to enhance the LLMs’ ToM performance within authentic social contexts.

\paragraph{Limited mental states types and evaluation modality} While our framework effectively models the interplay between belief, emotion, intention, and action based on established psychological theory \citep{d1995development}, there are opportunities to expand the scope of mental states evaluated. Future work could explore additional mental states, such as knowledge, to provide even richer insights into language models' ToM capabilities. While our dialogue-based evaluation approach has proven effective in assessing models' ToM abilities in dynamic contexts, future research could explore how these models perform in multimodal contexts that include visual and auditory cues. This extension would complement our text-based findings by examining how models track temporal changes in mental states across different modalities, though our current framework already provides robust insights into models' social reasoning capabilities.

\section*{Ethics Statement}

\paragraph{Annotators and contents}
We strictly adhere to the ACL Code of Ethics. We placed high importance on ensuring the comfort and well-being of our annotators. We advised them to stop the annotation process if they came across any information that caused them discomfort. We recruited annotators at a rate of 2 $\sim$ 3 times their local hourly minimum wage. We instruct the annotators to validate the data without bias and keep the content free from unsafe, toxic, biased, offensive, and harmful content. We utilize the models in accordance with their designated purpose. In summary, we make every effort to adhere to the ethical norms set forth by ACL.

\paragraph{Ethical Considerations.}
The theory of mind is a distinctive social cognitive capability that is intrinsic to humans. Assessing the Theory of Mind capacities of Large Language Models utilizing \textsc{DynToM} may result in anthropomorphic interpretations, attributing human-like qualities to LLMs. Nonetheless, it is imperative to clarify that our objective is not to anthropomorphize LLMs. Our objective is to evaluate the capacity of LLMs to comprehend and interpret human mental states, thus enhancing AI's interaction with humans in the social context.

\section*{Acknowledgements}
We would like to thank all reviewers for their insightful comments and suggestions to help improve the paper. This work was supported by the Research Grants Council of Hong Kong (GRF15213323, GRF 15209724 and TRS T43-518/24-N).

%% file: section/appendix.tex
\appendix

\section{The construction of the \textsc{DynToM}}\label{appd:sec:framework}
\subsection{the candidate pool of social location}
\label{candidate pool}
\begin{figure*}[h]
    \centering
    \includegraphics[width=\textwidth]{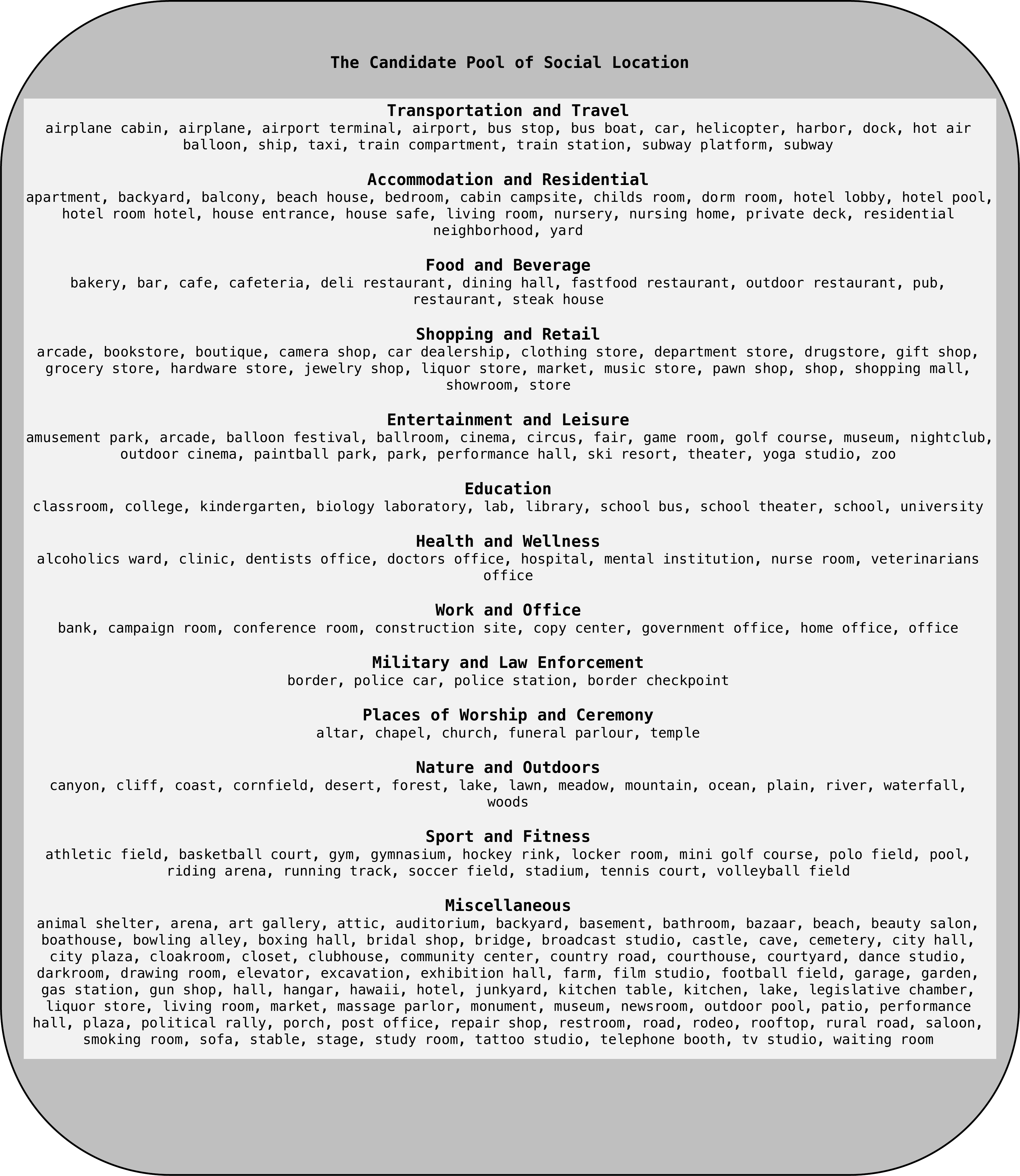}
    \caption{The candidate pool of social location.}
    \label{fig: social_setting_candidate}
\end{figure*}

The social location describes the environments where individuals live, work, and learn, which can significantly impact their mental states and behavior \citep{stokols1978environmental}. As shown in Figure \ref{fig: social_setting_candidate}, we have collected 13 types of social location types in total, adding up to 261 locations.

\subsection{the candidate pool of profile}
\label{append:social context}
\begin{figure*}[h]
    \centering
    \includegraphics[width=\textwidth]{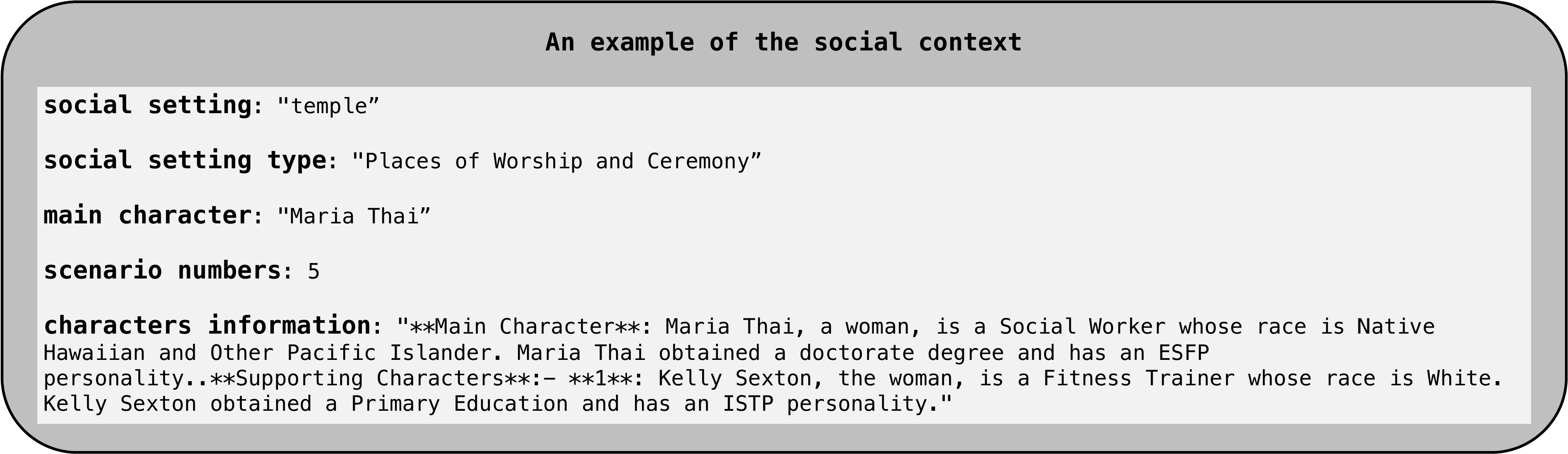}
    \caption{An example of the social context.}
    \label{fig: social_context}
\end{figure*}

\begin{figure*}[h]
    \centering
    \includegraphics[width=\textwidth]{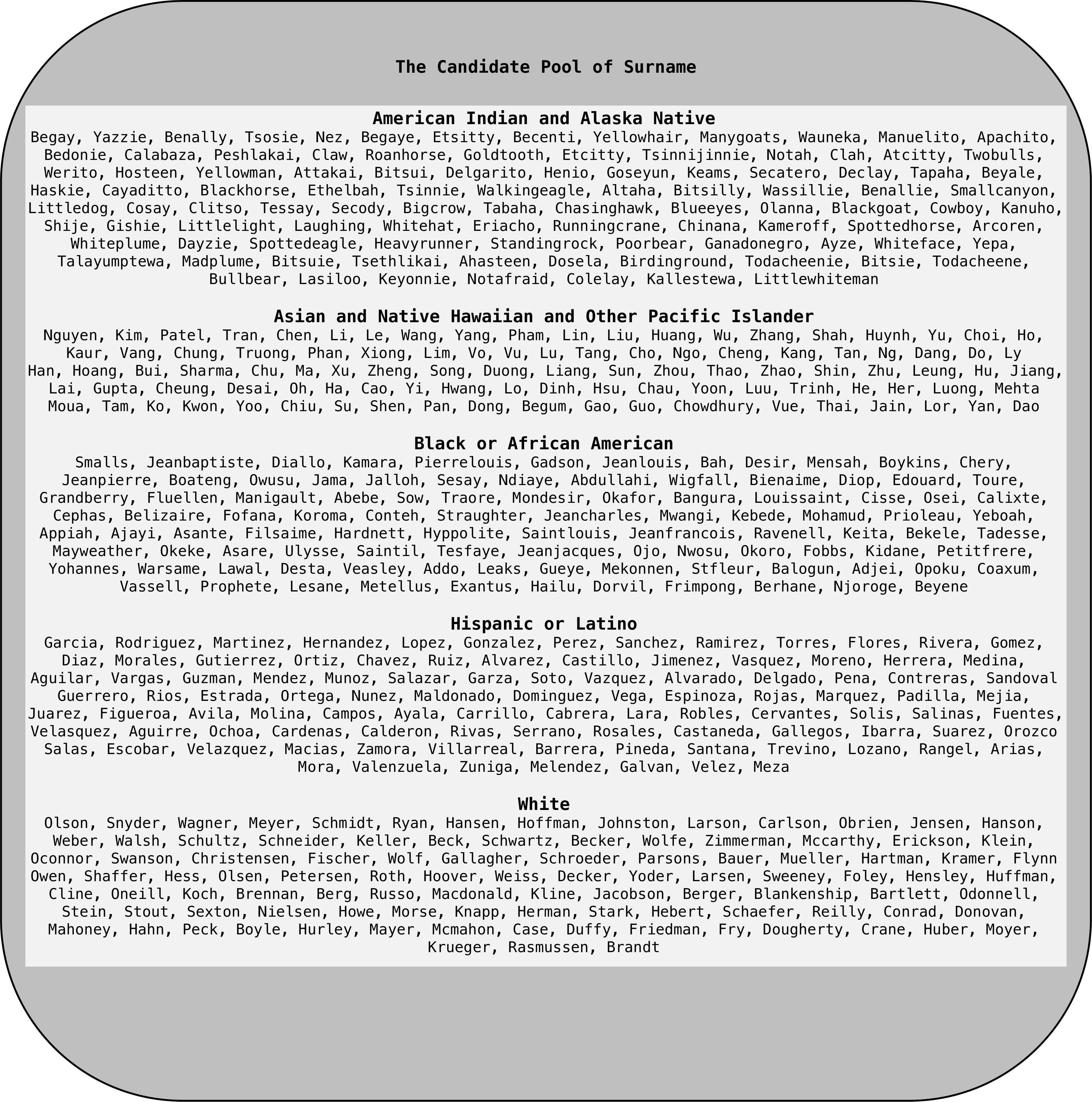}
    \caption{The races and their corresponding 100 most popular surnames.}
    \label{fig: surname_candidate}
\end{figure*}

\begin{figure*}[h]
    \centering
    \includegraphics[width=\textwidth]{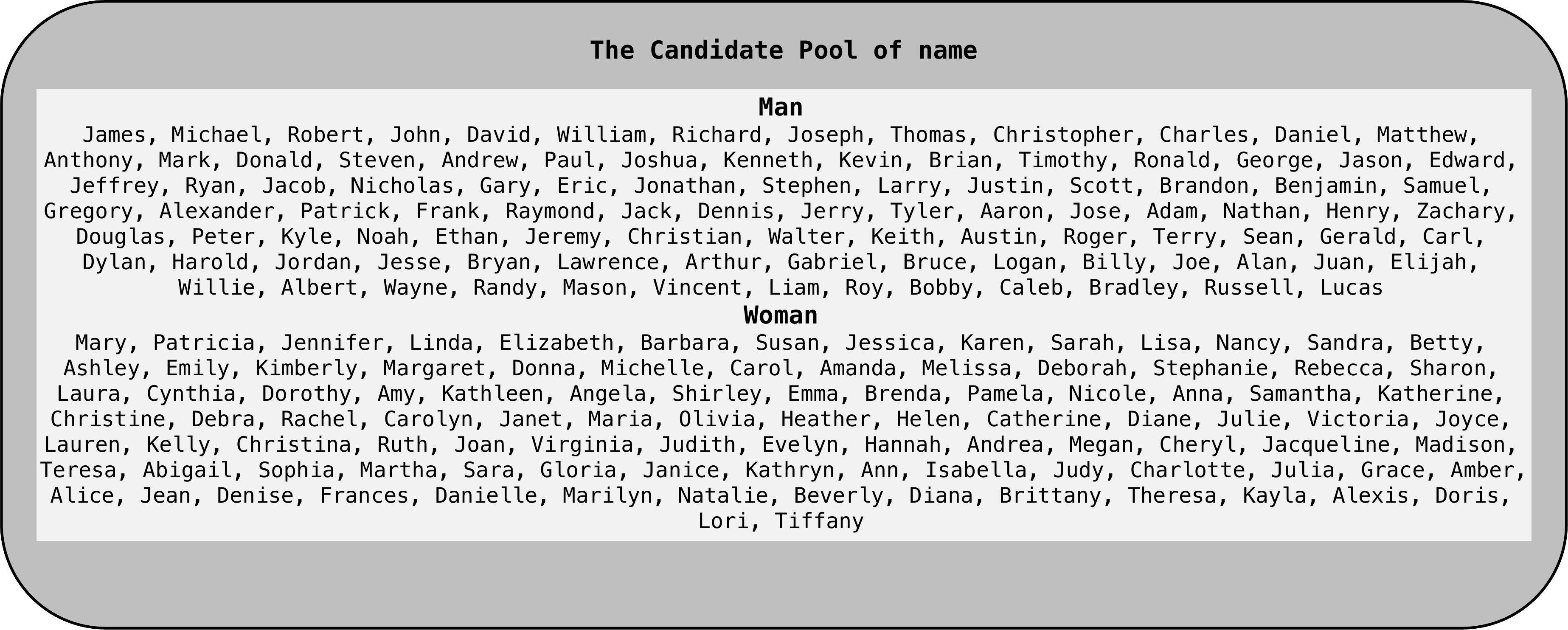}
    \caption{The genders and their corresponding 100 most popular names.}
    \label{fig: name_candidate}
\end{figure*}

\begin{figure*}[h]
    \centering
    \includegraphics[width=\textwidth]{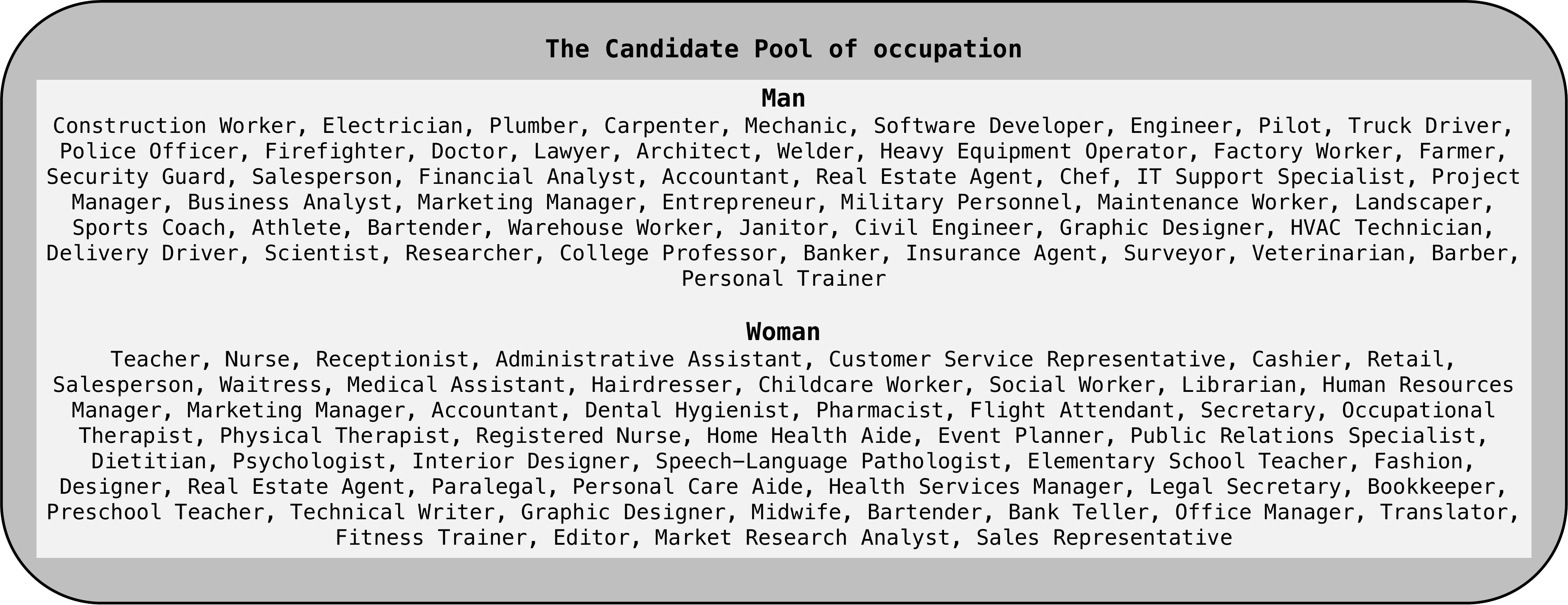}
    \caption{The genders and their corresponding 100 most popular occupations.}
    \label{fig: occupations_candidate}
\end{figure*}

\begin{figure*}[h]
    \centering
    \includegraphics[width=\textwidth]{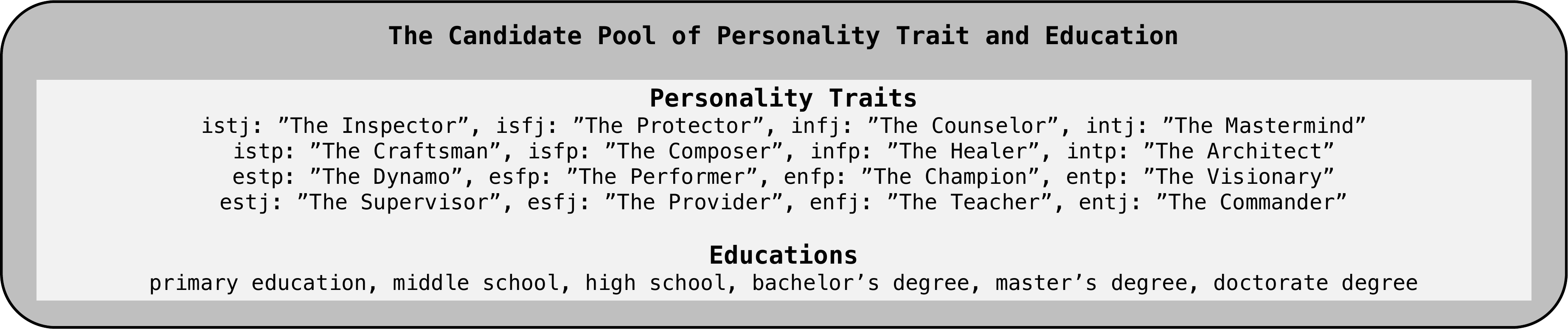}
    \caption{The personality traits and educations.}
    \label{fig: personality_education}
\end{figure*}

We conclude 7 aspects in the characters' profile: surname, name, gender, occupation, education, race, and personality traits. Their value can be found in Figure \ref{fig: surname_candidate}, \ref{fig: name_candidate}, \ref{fig: occupations_candidate}, and \ref{fig: personality_education}. The source of the surname, name, and occupation statistics are \href{https://www.census.gov/topics/population/genealogy/data/2010_surnames.html}{U.S. Census Bureau Homepage}, \href{https://www.ssa.gov/oact/babynames/decades/century.html}{The United States Social Security Administration}, and \href{https://www.bls.gov/oes/current/area_emp_chart/area_emp_chart.htm}{Bureau of Labor Statistics}, respectively. Figure \ref{fig: social_context} shows an example of the social background.

\subsection{The prompt used to generate the mental state trajectory }
\label{append:sketch of mental states}
\begin{figure*}[h]
    \centering
    \includegraphics[width=\textwidth]{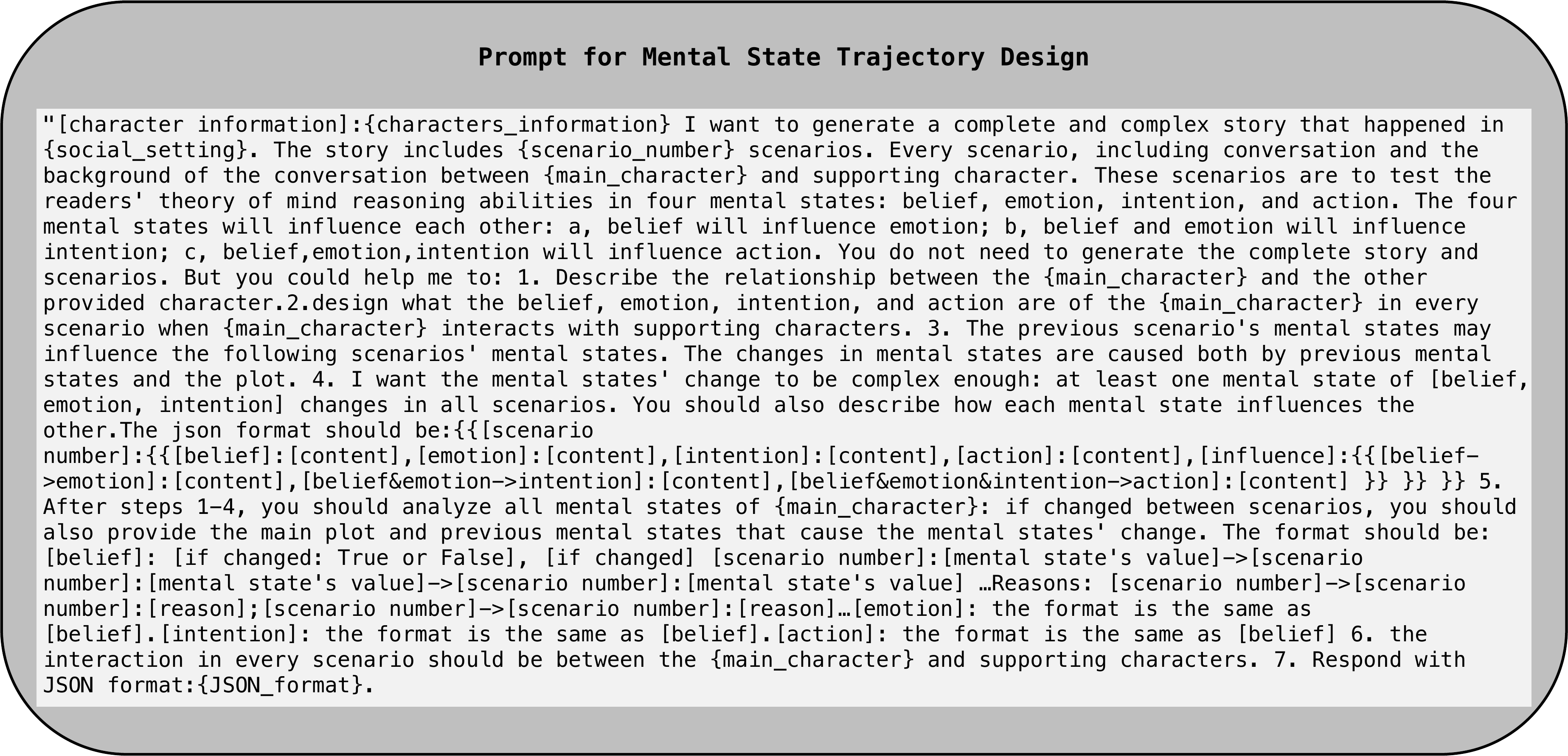}
    \caption{The prompt for the generation of the relationship between characters and the mental state trajectory.}
    \label{fig: prompt_sketch_mental_states}
\end{figure*}

\begin{figure*}[h]
\small
    \centering
    \includegraphics[width=0.7\textwidth]{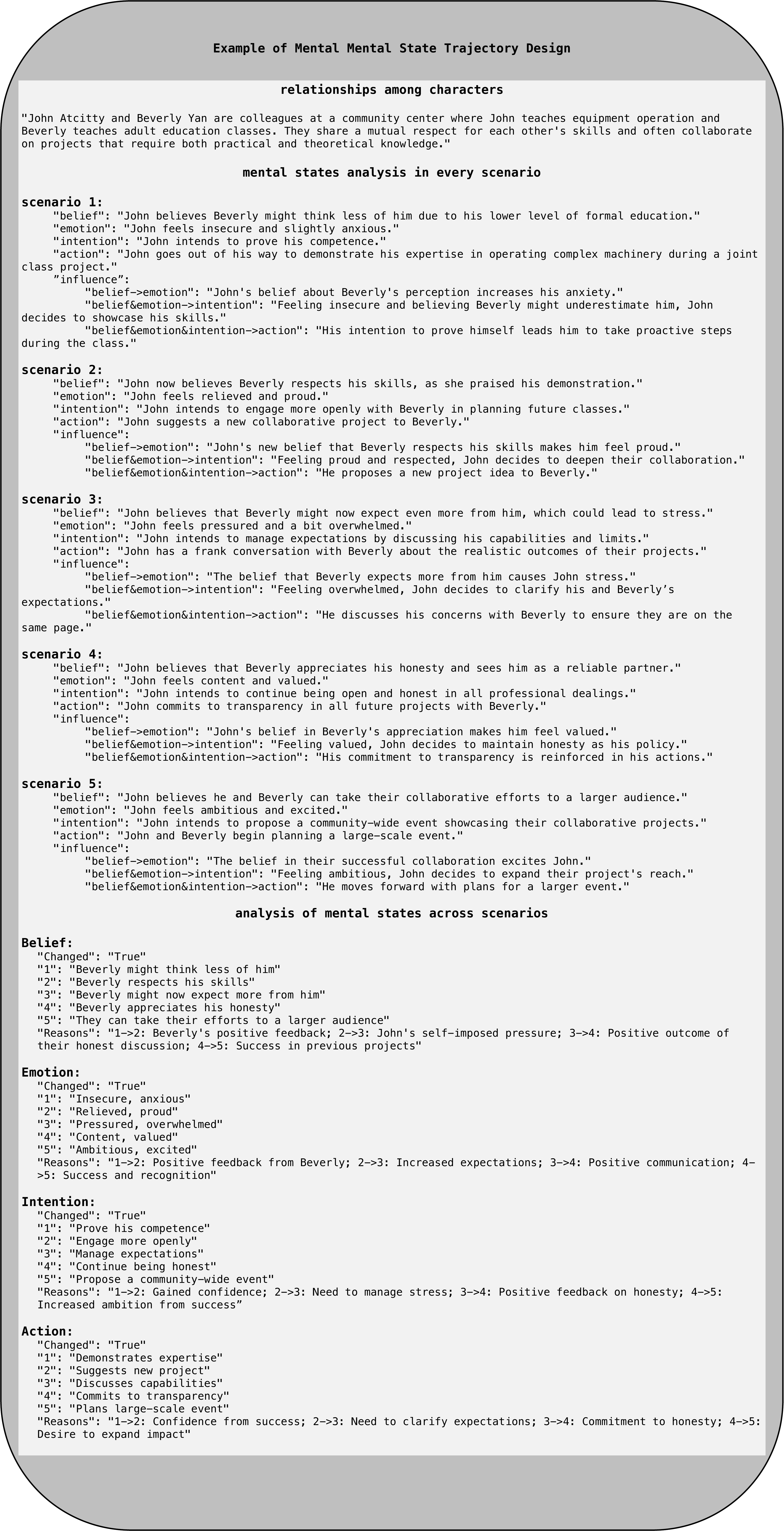}
    \caption{An example of the mental state trajectory.}
    \label{fig: sketch_mental_states}
\end{figure*}

As illustrated in Figure \ref{fig: prompt_sketch_mental_states}, the prompt is used to generate the mental state trajectory. In the holders of '\{\}' and '[]', the corresponding information will be input into this prompt. An example of the mental state trajectory is shown in Figure \ref{fig: sketch_mental_states}.

\subsection{The prompt used to generate the social scenarios}
\label{append:prompt of complete story}
\begin{figure*}[h]
    \centering
    \includegraphics[width=1\textwidth]{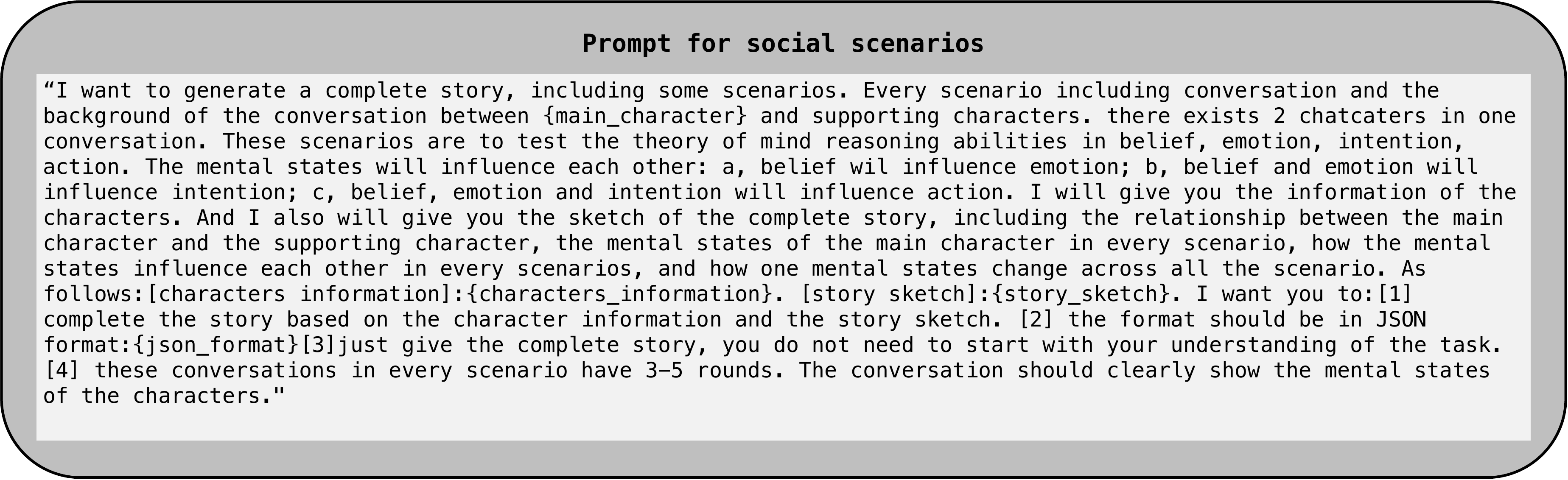}
    \caption{The prompt for the generation of the scenarios.}
    \label{fig: prompt_scenario}
\end{figure*}

\begin{figure*}[h]
    \centering
    \includegraphics[width=1\textwidth]{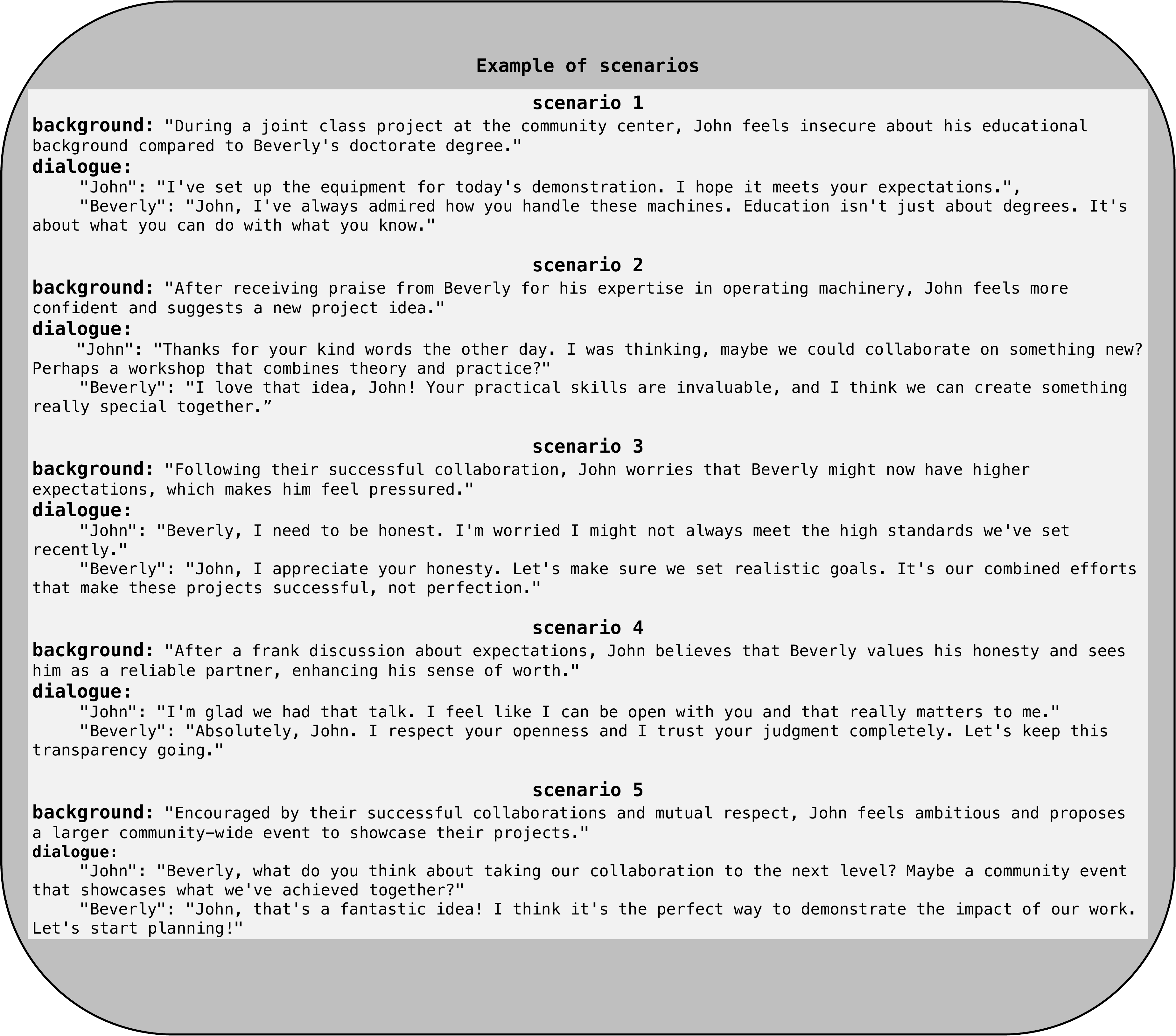}
    \caption{An example of the social scenarios.}
    \label{fig: scenario}
\end{figure*}

As illustrated in Figure \ref{fig: prompt_scenario}, the prompt is used to generate the social scenarios. In the holders of '\{\}' and '[]', the corresponding information will be input into this prompt. An example of the social scenario is shown in Figure \ref{fig: scenario}.

\subsection{The templates for the four types of questions and question examples}
\label{append:question template and examples}
\begin{figure*}[h]
    \centering
    \includegraphics[width=\textwidth]{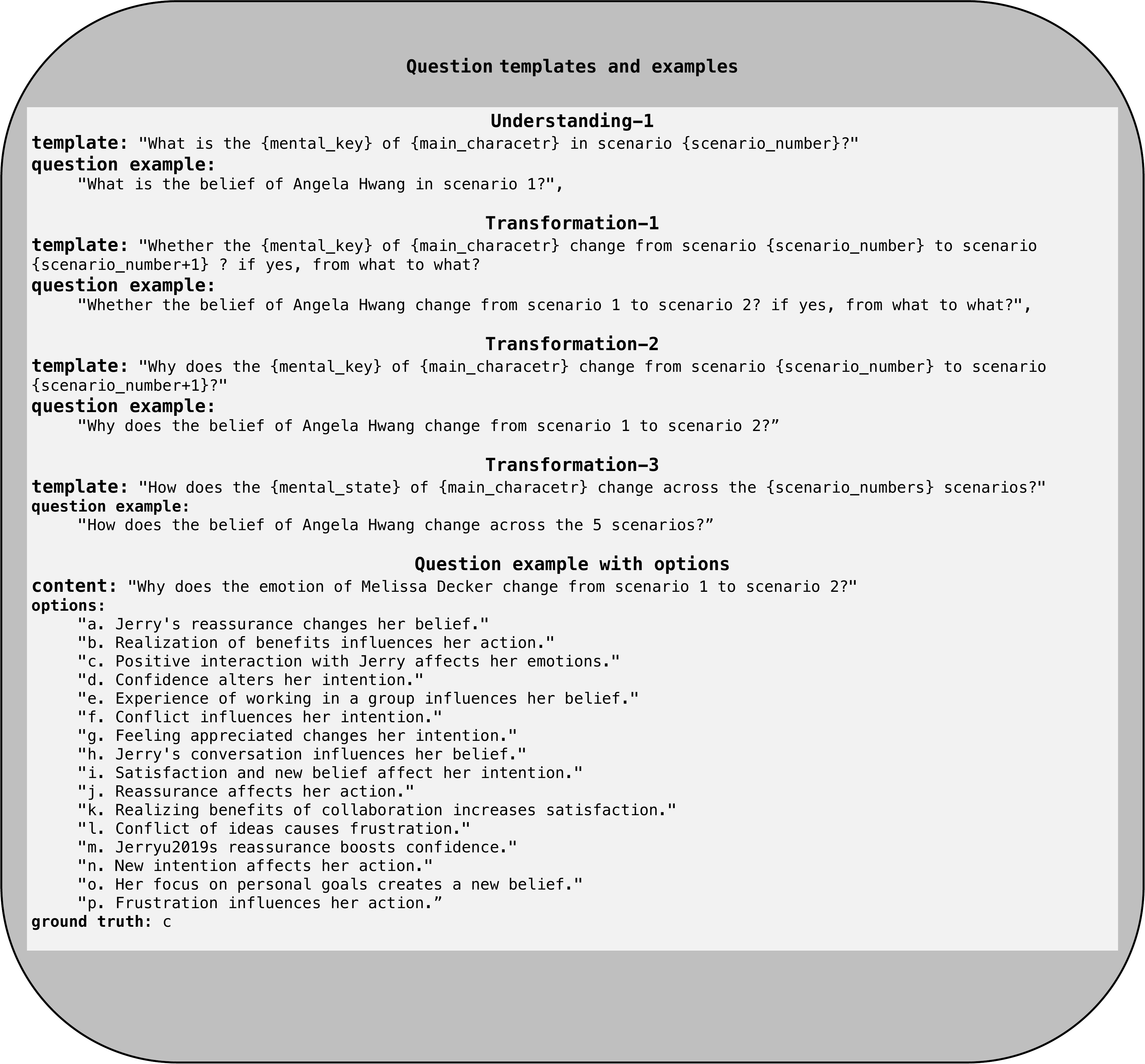}
    \caption{The examples of the types of questions and templates to generate these questions.}
    \label{fig: question_template}
\end{figure*}

We apply four predefined question templates to the social stage to generate questions, 71 questions for every social context in total. The four types of questions are: (1) (Understanding-1) What is the main character's ToM reasoning item in a specific scenario? (2) (Transformation-1) Does a ToM reasoning item change from scenario A to scenario B? (3) (Transformation-2) What causes a ToM reasoning item change from scenario A to scenario B? (4) (Transformation-3) How does the ToM reasoning item change across all the scenarios? The templates and the example of the four types of questions are shown in Figure \ref{fig: question_template}.

\subsection{Human validation of the Quality of \textsc{DynToM}}
\label{append:human evaluation}
\begin{figure*}[h]
    \centering
    \includegraphics[width=1\textwidth]{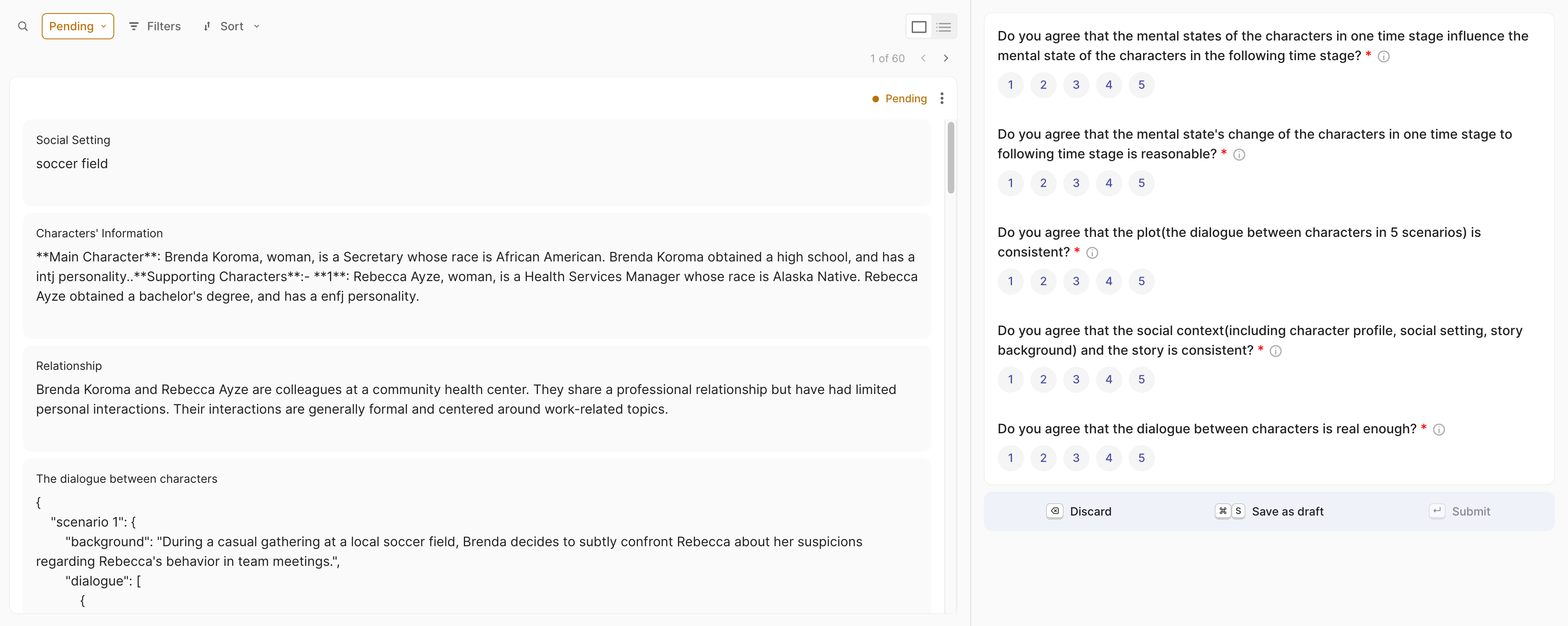}
    \caption{The platform to annotate the quality of the story.}
    \label{fig: data_annotation_benchmark_screen}
\end{figure*}

We apply \href{docs.argilla.io}{argilla} as the annotation platform. Figure \ref{fig: data_annotation_benchmark_screen} shows the annotation interface for data validation.

\section{Experiments}
\label{append:experiments}

\subsection{Model detail}
\label{append:model detail}
We evaluate a total of 10 popular LLMs, including GPT-4o, GPT-4-Turbo, Llama-3.1-8B, Llama-3.1-70B, Mistral-7B, Mixtral-8x7B, Qwen2-7B, Qwen2-72B, DeepSeek-V2, GLM-4. For all the LLMs, we strictly abide by their terms and get access through official APIs or model weights. Details about model versions, parameter sizes, context window sizes and the prompts used for the two methods are shown in Table \ref{tab:model detail}.

\begin{table}[!t]
\small
\centering
\adjustbox{width=0.5\textwidth}{
\begin{tabular}{lccc}
\toprule
Model & Version & Size & Context Length \\
\midrule
GPT-4o & \multicolumn{1}{r}{2024-05-13} & \textasciitilde  & 128k \\
GPT-4-Turbo & \multicolumn{1}{r}{2024-04-09} & \textasciitilde & 128k \\
Llama-3.1-8B & Instruct & 8B & 128k \\
Llama-3.1-70B & Instruct & 70B & 128k \\
Mistral-7B & Instruct-v0.3 & 7B & 32k \\
Mixtral-8x7B & Instruct-v0.1 & 8x7B & 32k \\
Qwen2-7B & Instruct & 7B & 128k \\
Qwen2-72B & Instruct & 72B & 128k \\
DeepSeek-V2 & Lite-Chat & 16B & 32k \\
GLM-4 & 9b-chat & 9B & 128k\\
\bottomrule
\end{tabular}}
\caption{The detail of models evaluated in our benchmark.}
\label{tab:model detail}
\end{table}

\subsection{Prompting methods}
\label{append:CoT_prompting}
\begin{figure*}[h]
    \centering
    \includegraphics[width=1\textwidth]{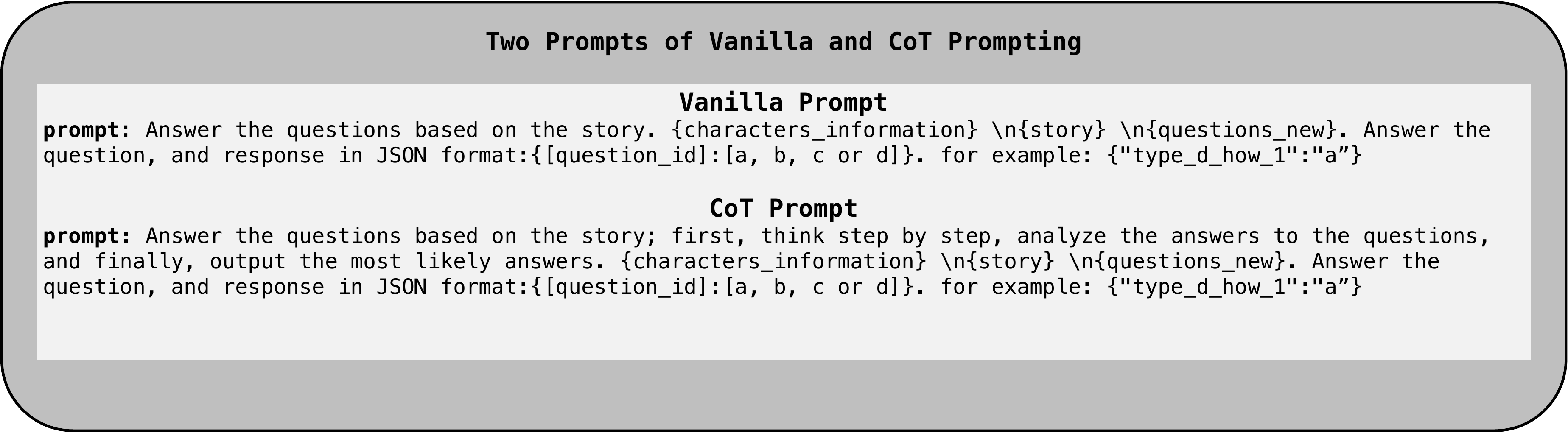}
    \caption{The prompts used for vanilla and CoT Prompting.}
    \label{fig: prompt_methods}
\end{figure*}

We employ two prompting methods: the vanilla prompting which directly asks LLMs to answer the questions, and the CoT prompting that elicits step-by-step reasoning before answering. The prompts used for the two methods are shown in Figure \ref{fig: prompt_methods}.

\subsection{Case Study for CoT Prompting}
\label{append:CoT_case_study}
\begin{figure*}[t]
    \small
    \centering
    \includegraphics[width=0.9\textwidth]{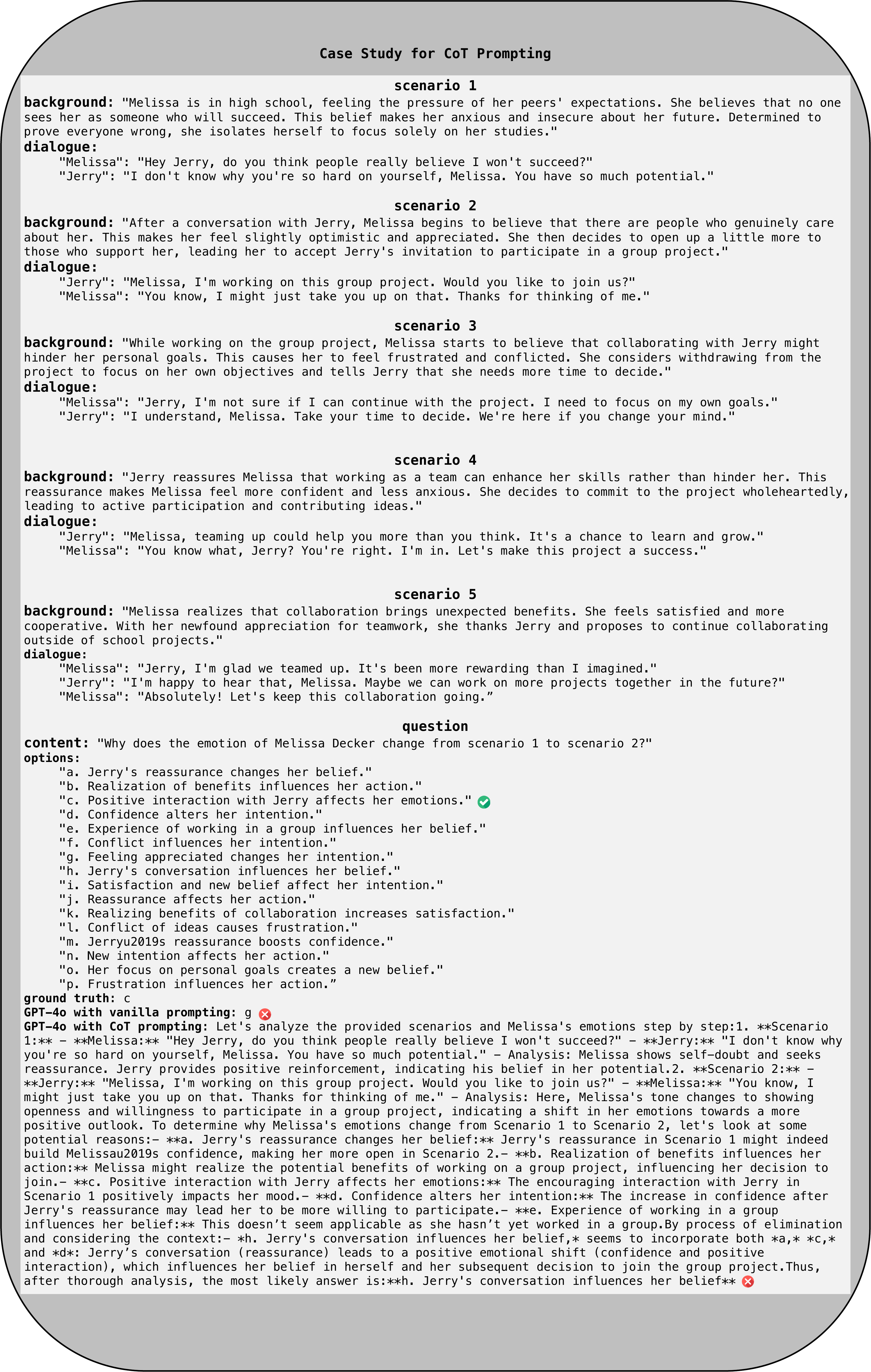}
    \caption{A case of CoT prompting on GPT-4o.}
    \label{fig: CoT_case_study}
\end{figure*}

Both ToM reasoning item and question-type results in Table \ref{tab:model_performance} indicate that CoT prompting doesn’t always improve LLMs’ ToM reasoning ability. We present a failure case of GPT-4o when CoT prompting is used in Figure \ref{fig: CoT_case_study}.

%% file: acl_latex.bbl
\begin{thebibliography}{40}
\providecommand{\natexlab}[1]{#1}

\bibitem[{Achiam et~al.(2023)Achiam, Adler, Agarwal, Ahmad, Akkaya, Aleman, Almeida, Altenschmidt, Altman, Anadkat et~al.}]{achiam2023gpt}
Josh Achiam, Steven Adler, Sandhini Agarwal, Lama Ahmad, Ilge Akkaya, Florencia~Leoni Aleman, Diogo Almeida, Janko Altenschmidt, Sam Altman, Shyamal Anadkat, et~al. 2023.
\newblock Gpt-4 technical report.
\newblock \emph{arXiv preprint arXiv:2303.08774}.

\bibitem[{Aher et~al.(2023)Aher, Arriaga, and Kalai}]{10.5555/3618408.3618425}
Gati Aher, Rosa~I. Arriaga, and Adam~Tauman Kalai. 2023.
\newblock Using large language models to simulate multiple humans and replicate human subject studies.
\newblock In \emph{Proceedings of the 40th International Conference on Machine Learning}, ICML'23. JMLR.org.

\bibitem[{Chen et~al.(2024)Chen, Wu, Zhou, Wen, Bi, Jiang, Cao, Hu, Lai, Xiong, and Huang}]{chen-etal-2024-tombench}
Zhuang Chen, Jincenzi Wu, Jinfeng Zhou, Bosi Wen, Guanqun Bi, Gongyao Jiang, Yaru Cao, Mengting Hu, Yunghwei Lai, Zexuan Xiong, and Minlie Huang. 2024.
\newblock \href {https://doi.org/10.18653/v1/2024.acl-long.847} {{T}o{MB}ench: Benchmarking theory of mind in large language models}.
\newblock In \emph{Proceedings of the 62nd Annual Meeting of the Association for Computational Linguistics (Volume 1: Long Papers)}, pages 15959--15983, Bangkok, Thailand. Association for Computational Linguistics.

\bibitem[{Dubey et~al.(2024)Dubey, Jauhri, Pandey, Kadian, Al-Dahle, Letman, Mathur, Schelten, Yang, Fan et~al.}]{dubey2024llama}
Abhimanyu Dubey, Abhinav Jauhri, Abhinav Pandey, Abhishek Kadian, Ahmad Al-Dahle, Aiesha Letman, Akhil Mathur, Alan Schelten, Amy Yang, Angela Fan, et~al. 2024.
\newblock The llama 3 herd of models.
\newblock \emph{arXiv preprint arXiv:2407.21783}.

\bibitem[{Dziri et~al.(2023)Dziri, Lu, Sclar, Li, Jiang, Lin, Welleck, West, Bhagavatula, Bras, Hwang, Sanyal, Ren, Ettinger, Harchaoui, and Choi}]{dziri2023faith}
Nouha Dziri, Ximing Lu, Melanie Sclar, Xiang~Lorraine Li, Liwei Jiang, Bill~Yuchen Lin, Sean Welleck, Peter West, Chandra Bhagavatula, Ronan~Le Bras, Jena~D. Hwang, Soumya Sanyal, Xiang Ren, Allyson Ettinger, Zaid Harchaoui, and Yejin Choi. 2023.
\newblock \href {https://openreview.net/forum?id=Fkckkr3ya8} {Faith and fate: Limits of transformers on compositionality}.
\newblock In \emph{Thirty-seventh Conference on Neural Information Processing Systems}.

\bibitem[{D’Andrade(1995)}]{d1995development}
Roy D’Andrade. 1995.
\newblock \emph{The development of cognitive anthropology}.
\newblock Cambridge University Press.

\bibitem[{Farrow et~al.(2017)Farrow, Grolleau, and Ibanez}]{farrow2017social}
Katherine Farrow, Gilles Grolleau, and Lisette Ibanez. 2017.
\newblock Social norms and pro-environmental behavior: A review of the evidence.
\newblock \emph{Ecological Economics}, 140:1--13.

\bibitem[{Gandhi et~al.(2023)Gandhi, Fr{\"a}nken, Gerstenberg, and Goodman}]{gandhi2023understanding}
Kanishk Gandhi, Jan-Philipp Fr{\"a}nken, Tobias Gerstenberg, and Noah Goodman. 2023.
\newblock \href {https://openreview.net/forum?id=8bqjirgxQM} {Understanding social reasoning in language models with language models}.
\newblock In \emph{Thirty-seventh Conference on Neural Information Processing Systems Datasets and Benchmarks Track}.

\bibitem[{Gandhi et~al.(2024)Gandhi, Fr{\"a}nken, Gerstenberg, and Goodman}]{gandhi2024understanding}
Kanishk Gandhi, Jan-Philipp Fr{\"a}nken, Tobias Gerstenberg, and Noah Goodman. 2024.
\newblock Understanding social reasoning in language models with language models.
\newblock \emph{Advances in Neural Information Processing Systems}, 36.

\bibitem[{GLM et~al.(2024)GLM, Zeng, Xu, Wang, Zhang, Yin, Zhang, Rojas, Feng, Zhao et~al.}]{glm2024chatglm}
Team GLM, Aohan Zeng, Bin Xu, Bowen Wang, Chenhui Zhang, Da~Yin, Dan Zhang, Diego Rojas, Guanyu Feng, Hanlin Zhao, et~al. 2024.
\newblock Chatglm: A family of large language models from glm-130b to glm-4 all tools.
\newblock \emph{arXiv preprint arXiv:2406.12793}.

\bibitem[{H\"{a}m\"{a}l\"{a}inen et~al.(2023)H\"{a}m\"{a}l\"{a}inen, Tavast, and Kunnari}]{10.1145/3544548.3580688}
Perttu H\"{a}m\"{a}l\"{a}inen, Mikke Tavast, and Anton Kunnari. 2023.
\newblock \href {https://doi.org/10.1145/3544548.3580688} {Evaluating large language models in generating synthetic hci research data: a case study}.
\newblock In \emph{Proceedings of the 2023 CHI Conference on Human Factors in Computing Systems}, CHI '23, New York, NY, USA. Association for Computing Machinery.

\bibitem[{Hua et~al.(2023)Hua, Fan, Li, Mei, Ji, Ge, Hemphill, and Zhang}]{hua2023war}
Wenyue Hua, Lizhou Fan, Lingyao Li, Kai Mei, Jianchao Ji, Yingqiang Ge, Libby Hemphill, and Yongfeng Zhang. 2023.
\newblock War and peace (waragent): Large language model-based multi-agent simulation of world wars.
\newblock \emph{arXiv preprint arXiv:2311.17227}.

\bibitem[{Jin et~al.(2024)Jin, Wu, Cao, Xiang, Kuo, Hu, Ullman, Torralba, Tenenbaum, and Shu}]{jin-etal-2024-mmtom}
Chuanyang Jin, Yutong Wu, Jing Cao, Jiannan Xiang, Yen-Ling Kuo, Zhiting Hu, Tomer Ullman, Antonio Torralba, Joshua Tenenbaum, and Tianmin Shu. 2024.
\newblock \href {https://doi.org/10.18653/v1/2024.acl-long.851} {{MMT}o{M}-{QA}: Multimodal theory of mind question answering}.
\newblock In \emph{Proceedings of the 62nd Annual Meeting of the Association for Computational Linguistics (Volume 1: Long Papers)}, pages 16077--16102, Bangkok, Thailand. Association for Computational Linguistics.

\bibitem[{Kim et~al.(2023)Kim, Sclar, Zhou, Bras, Kim, Choi, and Sap}]{kim-etal-2023-fantom}
Hyunwoo Kim, Melanie Sclar, Xuhui Zhou, Ronan Bras, Gunhee Kim, Yejin Choi, and Maarten Sap. 2023.
\newblock \href {https://doi.org/10.18653/v1/2023.emnlp-main.890} {{FANT}o{M}: A benchmark for stress-testing machine theory of mind in interactions}.
\newblock In \emph{Proceedings of the 2023 Conference on Empirical Methods in Natural Language Processing}, pages 14397--14413, Singapore. Association for Computational Linguistics.

\bibitem[{Le et~al.(2019)Le, Boureau, and Nickel}]{le-etal-2019-revisiting}
Matthew Le, Y-Lan Boureau, and Maximilian Nickel. 2019.
\newblock \href {https://doi.org/10.18653/v1/D19-1598} {Revisiting the evaluation of theory of mind through question answering}.
\newblock In \emph{Proceedings of the 2019 Conference on Empirical Methods in Natural Language Processing and the 9th International Joint Conference on Natural Language Processing (EMNLP-IJCNLP)}, pages 5872--5877, Hong Kong, China. Association for Computational Linguistics.

\bibitem[{Liu et~al.(2024{\natexlab{a}})Liu, Lin, Hewitt, Paranjape, Bevilacqua, Petroni, and Liang}]{liu-etal-2024-lost}
Nelson~F. Liu, Kevin Lin, John Hewitt, Ashwin Paranjape, Michele Bevilacqua, Fabio Petroni, and Percy Liang. 2024{\natexlab{a}}.
\newblock \href {https://doi.org/10.1162/tacl_a_00638} {Lost in the middle: How language models use long contexts}.
\newblock \emph{Transactions of the Association for Computational Linguistics}, 12:157--173.

\bibitem[{Liu et~al.(2024{\natexlab{b}})Liu, Zhao, Liu, Wang, and Peng}]{liu2024compeer}
Tianjian Liu, Hongzheng Zhao, Yuheng Liu, Xingbo Wang, and Zhenhui Peng. 2024{\natexlab{b}}.
\newblock Compeer: A generative conversational agent for proactive peer support.
\newblock \emph{arXiv preprint arXiv:2407.18064}.

\bibitem[{Nematzadeh et~al.(2018)Nematzadeh, Burns, Grant, Gopnik, and Griffiths}]{nematzadeh-etal-2018-evaluating}
Aida Nematzadeh, Kaylee Burns, Erin Grant, Alison Gopnik, and Tom Griffiths. 2018.
\newblock \href {https://doi.org/10.18653/v1/D18-1261} {Evaluating theory of mind in question answering}.
\newblock In \emph{Proceedings of the 2018 Conference on Empirical Methods in Natural Language Processing}, pages 2392--2400, Brussels, Belgium. Association for Computational Linguistics.

\bibitem[{Park et~al.(2023)Park, O'Brien, Cai, Morris, Liang, and Bernstein}]{park2023generative}
Joon~Sung Park, Joseph O'Brien, Carrie~Jun Cai, Meredith~Ringel Morris, Percy Liang, and Michael~S Bernstein. 2023.
\newblock Generative agents: Interactive simulacra of human behavior.
\newblock In \emph{Proceedings of the 36th annual acm symposium on user interface software and technology}, pages 1--22.

\bibitem[{Park et~al.(2022)Park, Popowski, Cai, Morris, Liang, and Bernstein}]{park2022social}
Joon~Sung Park, Lindsay Popowski, Carrie Cai, Meredith~Ringel Morris, Percy Liang, and Michael~S Bernstein. 2022.
\newblock Social simulacra: Creating populated prototypes for social computing systems.
\newblock In \emph{Proceedings of the 35th Annual ACM Symposium on User Interface Software and Technology}, pages 1--18.

\bibitem[{Premack and Woodruff(1978)}]{premack1978does}
David Premack and Guy Woodruff. 1978.
\newblock Does the chimpanzee have a theory of mind?
\newblock \emph{Behavioral and brain sciences}, 1(4):515--526.

\bibitem[{Sabour et~al.(2024)Sabour, Liu, Zhang, Liu, Zhou, Sunaryo, Lee, Mihalcea, and Huang}]{sabour-etal-2024-emobench}
Sahand Sabour, Siyang Liu, Zheyuan Zhang, June Liu, Jinfeng Zhou, Alvionna Sunaryo, Tatia Lee, Rada Mihalcea, and Minlie Huang. 2024.
\newblock \href {https://doi.org/10.18653/v1/2024.acl-long.326} {{E}mo{B}ench: Evaluating the emotional intelligence of large language models}.
\newblock In \emph{Proceedings of the 62nd Annual Meeting of the Association for Computational Linguistics (Volume 1: Long Papers)}, pages 5986--6004, Bangkok, Thailand. Association for Computational Linguistics.

\bibitem[{Sap et~al.(2022)Sap, Le~Bras, Fried, and Choi}]{sap-etal-2022-neural}
Maarten Sap, Ronan Le~Bras, Daniel Fried, and Yejin Choi. 2022.
\newblock \href {https://doi.org/10.18653/v1/2022.emnlp-main.248} {Neural theory-of-mind? on the limits of social intelligence in large {LM}s}.
\newblock In \emph{Proceedings of the 2022 Conference on Empirical Methods in Natural Language Processing}, pages 3762--3780, Abu Dhabi, United Arab Emirates. Association for Computational Linguistics.

\bibitem[{Sclar et~al.(2023)Sclar, Kumar, West, Suhr, Choi, and Tsvetkov}]{sclar-etal-2023-minding}
Melanie Sclar, Sachin Kumar, Peter West, Alane Suhr, Yejin Choi, and Yulia Tsvetkov. 2023.
\newblock \href {https://doi.org/10.18653/v1/2023.acl-long.780} {Minding language models' (lack of) theory of mind: A plug-and-play multi-character belief tracker}.
\newblock In \emph{Proceedings of the 61st Annual Meeting of the Association for Computational Linguistics (Volume 1: Long Papers)}, pages 13960--13980, Toronto, Canada. Association for Computational Linguistics.

\bibitem[{Shapira et~al.(2024)Shapira, Levy, Alavi, Zhou, Choi, Goldberg, Sap, and Shwartz}]{shapira-etal-2024-clever}
Natalie Shapira, Mosh Levy, Seyed~Hossein Alavi, Xuhui Zhou, Yejin Choi, Yoav Goldberg, Maarten Sap, and Vered Shwartz. 2024.
\newblock \href {https://aclanthology.org/2024.eacl-long.138/} {Clever hans or neural theory of mind? stress testing social reasoning in large language models}.
\newblock In \emph{Proceedings of the 18th Conference of the European Chapter of the Association for Computational Linguistics (Volume 1: Long Papers)}, pages 2257--2273, St. Julian{'}s, Malta. Association for Computational Linguistics.

\bibitem[{Shi et~al.(2024)Shi, Ye, Fang, Jin, Isik, Kuo, and Shu}]{shi2024mumatommultimodalmultiagenttheory}
Haojun Shi, Suyu Ye, Xinyu Fang, Chuanyang Jin, Leyla Isik, Yen-Ling Kuo, and Tianmin Shu. 2024.
\newblock \href {https://arxiv.org/abs/2408.12574} {Muma-tom: Multi-modal multi-agent theory of mind}.
\newblock \emph{Preprint}, arXiv:2408.12574.

\bibitem[{Stokols(1978)}]{stokols1978environmental}
Daniel Stokols. 1978.
\newblock Environmental psychology.

\bibitem[{Turner(1988)}]{turner1988theory}
Jonathan~H Turner. 1988.
\newblock \emph{A Theory of Social Interaction}.
\newblock Stanford University Press.

\bibitem[{Ullman(2023)}]{ullman2023large}
Tomer Ullman. 2023.
\newblock Large language models fail on trivial alterations to theory-of-mind tasks.
\newblock \emph{arXiv preprint arXiv:2302.08399}.

\bibitem[{Wang et~al.(2024)Wang, Xiao, Li, Song, Xu, Tan, and Li}]{wang2024towards}
Jiashuo Wang, Yang Xiao, Yanran Li, Changhe Song, Chunpu Xu, Chenhao Tan, and Wenjie Li. 2024.
\newblock Towards a client-centered assessment of llm therapists by client simulation.
\newblock \emph{arXiv preprint arXiv:2406.12266}.

\bibitem[{Wang et~al.(2023)Wang, Wei, Schuurmans, Le, Chi, Narang, Chowdhery, and Zhou}]{wang2023selfconsistency}
Xuezhi Wang, Jason Wei, Dale Schuurmans, Quoc~V Le, Ed~H. Chi, Sharan Narang, Aakanksha Chowdhery, and Denny Zhou. 2023.
\newblock \href {https://openreview.net/forum?id=1PL1NIMMrw} {Self-consistency improves chain of thought reasoning in language models}.
\newblock In \emph{The Eleventh International Conference on Learning Representations}.

\bibitem[{Wei et~al.(2022)Wei, Wang, Schuurmans, Bosma, Xia, Chi, Le, Zhou et~al.}]{wei2022chain}
Jason Wei, Xuezhi Wang, Dale Schuurmans, Maarten Bosma, Fei Xia, Ed~Chi, Quoc~V Le, Denny Zhou, et~al. 2022.
\newblock Chain-of-thought prompting elicits reasoning in large language models.
\newblock \emph{Advances in neural information processing systems}, 35:24824--24837.

\bibitem[{Wilf et~al.(2024)Wilf, Lee, Liang, and Morency}]{wilf-etal-2024-think}
Alex Wilf, Sihyun Lee, Paul~Pu Liang, and Louis-Philippe Morency. 2024.
\newblock \href {https://doi.org/10.18653/v1/2024.acl-long.451} {Think twice: Perspective-taking improves large language models' theory-of-mind capabilities}.
\newblock In \emph{Proceedings of the 62nd Annual Meeting of the Association for Computational Linguistics (Volume 1: Long Papers)}, pages 8292--8308, Bangkok, Thailand. Association for Computational Linguistics.

\bibitem[{Wu et~al.(2023)Wu, He, Jia, Mihalcea, Chen, and Deng}]{wu-etal-2023-hi}
Yufan Wu, Yinghui He, Yilin Jia, Rada Mihalcea, Yulong Chen, and Naihao Deng. 2023.
\newblock \href {https://doi.org/10.18653/v1/2023.findings-emnlp.717} {Hi-{T}o{M}: A benchmark for evaluating higher-order theory of mind reasoning in large language models}.
\newblock In \emph{Findings of the Association for Computational Linguistics: EMNLP 2023}, pages 10691--10706, Singapore. Association for Computational Linguistics.

\bibitem[{Xiao et~al.(2023)Xiao, Cheng, Fu, Wang, Li, and Liu}]{xiao2023far}
Yang Xiao, Yi~Cheng, Jinlan Fu, Jiashuo Wang, Wenjie Li, and Pengfei Liu. 2023.
\newblock How far are we from believable ai agents? a framework for evaluating the believability of human behavior simulation.
\newblock \emph{arXiv preprint arXiv:2312.17115}.

\bibitem[{Xie et~al.(2024)Xie, Chen, Jia, Ye, Lai, Shu, Gu, Bibi, Hu, Jurgens, Evans, Torr, Ghanem, and Li}]{xie2024can}
Chengxing Xie, Canyu Chen, Feiran Jia, Ziyu Ye, Shiyang Lai, Kai Shu, Jindong Gu, Adel Bibi, Ziniu Hu, David Jurgens, James Evans, Philip Torr, Bernard Ghanem, and Guohao Li. 2024.
\newblock \href {https://openreview.net/forum?id=CeOwahuQic} {Can large language model agents simulate human trust behavior?}
\newblock In \emph{The Thirty-eighth Annual Conference on Neural Information Processing Systems}.

\bibitem[{Xu et~al.(2024)Xu, Zhao, Zhu, Du, and He}]{xu-etal-2024-opentom}
Hainiu Xu, Runcong Zhao, Lixing Zhu, Jinhua Du, and Yulan He. 2024.
\newblock \href {https://doi.org/10.18653/v1/2024.acl-long.466} {{O}pen{T}o{M}: A comprehensive benchmark for evaluating theory-of-mind reasoning capabilities of large language models}.
\newblock In \emph{Proceedings of the 62nd Annual Meeting of the Association for Computational Linguistics (Volume 1: Long Papers)}, pages 8593--8623, Bangkok, Thailand. Association for Computational Linguistics.

\bibitem[{Yang et~al.(2024)Yang, Yang, Zhang, Hui, Zheng, Yu, Li, Liu, Huang, Wei et~al.}]{yang2024qwen2}
An~Yang, Baosong Yang, Beichen Zhang, Binyuan Hui, Bo~Zheng, Bowen Yu, Chengyuan Li, Dayiheng Liu, Fei Huang, Haoran Wei, et~al. 2024.
\newblock Qwen2. 5 technical report.
\newblock \emph{arXiv preprint arXiv:2412.15115}.

\bibitem[{Yoon et~al.(2024)Yoon, He, Echterhoff, and McAuley}]{yoon-etal-2024-evaluating}
Se-eun Yoon, Zhankui He, Jessica Echterhoff, and Julian McAuley. 2024.
\newblock \href {https://doi.org/10.18653/v1/2024.naacl-long.83} {Evaluating large language models as generative user simulators for conversational recommendation}.
\newblock In \emph{Proceedings of the 2024 Conference of the North American Chapter of the Association for Computational Linguistics: Human Language Technologies (Volume 1: Long Papers)}, pages 1490--1504, Mexico City, Mexico. Association for Computational Linguistics.

\bibitem[{Ziems et~al.(2023)Ziems, Dwivedi-Yu, Wang, Halevy, and Yang}]{ziems-etal-2023-normbank}
Caleb Ziems, Jane Dwivedi-Yu, Yi-Chia Wang, Alon Halevy, and Diyi Yang. 2023.
\newblock \href {https://doi.org/10.18653/v1/2023.acl-long.429} {{N}orm{B}ank: A knowledge bank of situational social norms}.
\newblock In \emph{Proceedings of the 61st Annual Meeting of the Association for Computational Linguistics (Volume 1: Long Papers)}, pages 7756--7776, Toronto, Canada. Association for Computational Linguistics.

\end{thebibliography}
